\documentclass[sigconf,nonacm]{acmart}
\AtBeginDocument{%
  }

\setcopyright{acmlicensed}
\copyrightyear{2018}
\acmYear{2018}
\acmDOI{XXXXXXX.XXXXXXX}
\acmConference[Conference acronym 'XX]{Make sure to enter the correct
  conference title from your rights confirmation email}{June 03--05,
  2018}{Woodstock, NY}
\acmISBN{978-1-4503-XXXX-X/2018/06}

\usepackage[printonlyused,nolist,nohyperlinks]{acronym}
\usepackage{pdfpages}
\begin{document}

\acrodef{ML}{machine learning}
\acrodef{FMEA}{Failure Mode and Effects Analysis}
\acrodef{HAZOP}{Hazard and Operability Analysis}
\acrodef{FTA}{Fault Tree Analysis}
\acrodef{STPA}{System Theoretic Process Analysis}
\acrodef{T2I}{Text-to-image}
\acrodef{UCA}{Unsafe Control Action}
\acrodef{AI}{Artificial Intelligence}
\acrodef{FATE}{Fairness, Accountability, Transparency and Ethics}
\acrodef{HCI}{Human-Computer Interaction}
\acrodef{RAI}{Responsible AI}
\acrodef{RL}{Reinforcement Learning}
\acrodef{RAISA}{Responsible AI System Analysis}
\acrodef{SHAPE-AI}{System Hazard Analysis and Process Evaluation for AI}
\acrodef{PHASE}{Process-oriented Hazard Analysis for AI Systems}

\title{From Silos to Systems: Process-Oriented Hazard Analysis for AI Systems}

\author{Shalaleh Rismani}
\authornote{Corresponding author.}
\email{shalaleh.rismani@mail.mcgill.ca}
\affiliation{%
  \institution{McGill University}
  \city{Montreal}
  \state{Quebec}
  \country{Canada}
}

\author{Roel Dobbe}
\email{R.I.J.Dobbe@tudelft.nl}
\affiliation{%
  \institution{Delft University of Technology}
  \city{Delft}
  \country{The Netherlands}
}

\author{AJung Moon}
\email{ajung.moon@mcgill.ca}
\affiliation{%
  \institution{McGill University}
  \city{Montreal}
  \state{Quebec}
  \country{Canada}
}

\renewcommand{\shortauthors}{Rismani et al.}

\begin{abstract}
To effectively address potential harms from \acf{AI} systems, it is essential to identify and mitigate system-level hazards. 
Current analysis approaches focus on individual components of an \ac{AI} system, like training data or models, in isolation, overlooking hazards from component interactions or how they are situated within a company’s development process. To this end, we draw from the established field of system safety, which considers safety as an emergent property of the entire system, not just its components. In this work, we translate \acf{STPA} -- a recognized system safety framework for analyzing \ac{AI} development and operation processes. We focus on systems that rely on machine learning algorithms and conduct \ac{STPA} on three case studies involving linear regression, reinforcement learning, and transformer-based generative models. Our analysis explored how \ac{STPA}'s control and system-theoretic perspectives apply to \ac{AI} systems and whether unique \ac{AI} traits—such as model opacity, capability uncertainty, and output complexity—necessitate significant modifications to the framework.
We find that the key concepts and steps of conducting an \ac{STPA} apply to \ac{AI} systems but require targeted adaptations to address \ac{AI}-specific challenges---such as model opacity, autonomy, and output complexity---that arise to differing degrees across \ac{AI} paradigms such as linear models, reinforcement learning, and generative models. We present the \acf{PHASE} as a guideline that adapts \ac{STPA} concepts for \ac{AI}. Applying and interpreting \ac{STPA} using the \ac{PHASE} guidelines 
enables four key affordances for analysts responsible for managing \ac{AI} system harms: 1) detection of hazards at the systems level, including those from accumulation of disparate issues; 2) explicit acknowledgment of social factors contributing to experiences of algorithmic harms; 3) creation of traceable accountability chains between harms and those who can mitigate the harm; and 4) ongoing monitoring and mitigation of new hazards.
\end{abstract}




\maketitle

\section{Introduction}
\label{introduction}

When it comes to safety and risk management of \acf{AI}-based technologies, there is an emerging recognition that hazards - potential sources of harm - from such technologies need to be viewed and analyzed at the (sociotechnical) system level \cite{Dobbe2022-ql, Weidinger2023-pe}. Regulatory efforts such as the European Union \ac{AI} Act explicitly distinguish between the terms \ac{AI} model and \ac{AI} system \cite{EUAIActArticle3}. In parallel, major \ac{AI} developers are releasing system cards in addition to model and data cards, recognizing that the performance of many recently released Application Programming Interfaces (APIs) is the result of the amalgamation of multiple models and software components\cite{meta-system-card,GPTsystemcard}. Additionally, the role of the algorithmic supply chain, which includes the actors involved in developing, integrating, and using an AI system, is increasingly recognized as central to understanding how harms emerge from \ac{AI} systems \cite{Cobbe2023-eg, Attard-Frost2025-tv}. Finally, there is growing recognition of the importance of defense-in-depth approaches to mitigating risks from \ac{AI} systems, which advocate for multiple mitigation mechanisms applied across the \ac{AI} development lifecycle \cite{Bengio2025InternationalAISafetyReport}. Despite growing views of AI-based technologies as sociotechnical systems, there has been surprisingly minimal empirical effort to develop and examine frameworks that facilitate hazard analysis at the system level for AI systems. In this work, we draw on a well-established hazard analysis framework from the field of system safety and examine how it could be applied to analyze the processes involved in developing and integrating an AI system. 

Hazard analysis has a long-standing history in the field of safety engineering, and a track record of being an essential component of developing and deploying complex safety-critical technologies such as nuclear power plants and passenger planes \cite{Carlson2012-iq, Knight2002-dy, Leveson2011-fo}. 
Yet, the \ac{AI} community has been slow at embracing the fundamental know-how and practices well established in safety engineering \cite{Carlson2012-iq, Knight2002-dy, Leveson2011-fo}. 
We specifically draw on the field of system safety, one of the most recent and significant developments in safety engineering\cite{Leveson2004-jn}. System safety emerged from the observation that safe systems cannot be achieved by merely analyzing technical components in isolation. Instead, system safety emphasizes the necessity of considering the broader social and technological context. System safety frameworks provide the means of analyzing systems as a whole rather than conducting multiple, separate analyses of their composing sub-systems (e.g., \ac{ML} model, development team, user) \cite{Nabavi2023-ce, Leveson2017-di}.

In this work we examine to what degree a well-established hazard analysis framework can be readily applied to assess AI systems and what it affords to those conducting the assessment. In particular, we focus on \ac{STPA}, a framework well-known for its utility in analyzing complex technical systems \cite{Leveson_undated-zz}. \ac{STPA} offers two powerful affordances essential for examining hazards at the system level: 1) It provides the means to map out and analyze complex social and technical systems, leveraging concepts from system theory and control theory. 2) It centers on the potential loss of stakeholder values, which facilitates harm analysis beyond safety-critical harms (i.e., loss of life, damage to property, and environmental harm). However, many of the current guidelines for conducting \ac{STPA} primarily focus on traditionally safety-critical systems such as nuclear power plants.
There are currently only a handful of examples of its application to AI systems and algorithmic harms.\cite{Jatho2023-zr, Rismani2023-xt}. When applying STPA to AI systems, we build on existing work while focusing on characteristics and challenges unique to conducting hazard analysis for \ac{AI}, such as system opacity, capability uncertainty, and output complexity, as well as the diverse range of potential harms that could result from their adoption. Crucially, these characteristics are not uniform across \ac{AI} systems: a linear regression model, a reinforcement learning agent, and a transformer-based generative model differ substantially in their opacity, autonomy, and output complexity, and therefore place different demands on each stage of the analysis. We treat this heterogeneity as central rather than incidental to our study, and our case studies are chosen to expose it. 
Specifically, we ask the following research questions:
\begin{itemize}
    \item RQ1: How can the steps of the \ac{STPA} framework be translated to analyze potential hazards from AI systems? 
    \item RQ2: What does \ac{STPA} afford an analyst in identifying, assessing and mitigating potential harms?
\end{itemize}

We conducted three case studies spanning three \ac{AI} systems (linear regression, reinforcement learning, transformer-based text-to-image models) across two domains of application (creative practice and medicine). Following the existing handbook by Leveson and Thomas, we conducted \ac{STPA} on the three case studies \cite{Leveson_undated-zz} and documented our interpretive and reflexive experience in the process. We find that \ac{STPA}-based analysis of AI provides the means for detecting potential sources of harm along both the development and operation of AI systems.
We present a guideline, \acf{PHASE}, documenting our translation of \ac{STPA} for \ac{AI}, considering the aforementioned unique characteristics. The guideline introduces key considerations that need to be accounted for when applying \ac{STPA} for \ac{AI} systems. 
This work contributes to the growing body of responsible and safe \ac{AI} practices discourse by: 1) producing \ac{PHASE} guidelines that adapt \ac{STPA} to the specific challenges of \ac{AI} systems and enable system-level hazard analysis for \ac{AI}; 2) providing three diverse case studies to illustrate the outcome of the analyses, and 3) presenting four affordances as shown by empirical investigation of these case studies.


\section{Background}
\label{background}
In this section, we situate our work in today's \ac{AI} safety discourse, \ac{RAI}, and system safety literature. We then briefly describe the key tenets and the steps of the \ac{STPA} framework in Section \ref{STPA_steps}. 

Note that in this work, we use the Organisation for Economic Co-operation and Development (OECD) definition of an \ac{AI} system: \textit{``a machine-based system that, for explicit or implicit objectives, infers, from the input it receives, how to generate outputs such as predictions, content, recommendations, or decisions that can influence physical or virtual environments."} \cite{Oecd2024-kr}. An AI system consists of multiple components, including the input dataset, model, and resulting output. 

\subsection{Examining Potential Sources of Algorithmic Harm}
Alongside the increased public awareness of how \ac{AI} systems can introduce societal disruptions and harm to individuals, \ac{RAI} scholars have been actively studying the various types of algorithmic harms and means of harm mitigation. \citet{Shelby2023-to}, for example, present at least 170 articles related to sociotechnical harms of algorithmic systems -- defined as ``adverse lived experiences of individuals and communities.''
The rapid advancements and adoption of large-scale, multi-modal, and general-purpose \ac{ML} models in the last five years 
have further intensified the discussions about \ac{AI}-related harms \cite{Bender2021-wy, Birhane2021-ry, Solaiman2023-dp}. 
The emerging epistemic community of \ac{AI} safety is also drawing more attention to concerns about \ac{AI}-driven existential and catastrophic risks to humanity \cite{Ahmed2023-kv, Bucknall2022-hw,Russell2022-hc}. 
To date, the types of harm that can be expected from current and near-future \ac{AI} systems remain a subject of heated debate.

Alongside these discussions are multitudes of theoretical frameworks, tools, and processes proposed to analyze and mitigate potential sources of such harms.
Qualitative and normative frameworks are commonly used to identify societal, ethical, and legal factors, helping analysts reason about how AI systems may produce unfair, exclusionary, or otherwise harmful outcomes in specific contexts of use \cite{Watkins2021-wc, Mantelero2021-cr, Reisman2018-fs}. Complementing these approaches, technical tools, often implemented as software libraries or code bases, are employed to examine potential sources of harm within datasets and models, such as data quality issues, bias, and performance disparities\cite{Barocas2021-sr,ODSC-Open_Data_Science2022-re}.
Safety and security engineering tools -- inspired by processes such as red teaming \cite{Ganguli2022-qn} or \ac{FMEA} -- aim to anticipate how AI models may behave under adversarial, unexpected, or edge-case conditions, thereby identifying model-level failure modes that could lead to downstream harms \cite{Rismani2021-gr, Raji2020-aq, Li2022-wj,Gyevnar2025-uw};
Additionally, participatory and community-based research methods have been adopted to uncover harms that emerge through real-world interactions between \ac{AI} systems, individuals, and communities, particularly those that may be overlooked by purely technical evaluations \cite{Lima2023-bs, Blodgett2022-db, Wenzel2023-pi, Bai2022-ai}. 

Existing methods are designed to be effective in analyzing hazards either by focusing on a single component of an \ac{AI} system (e.g., representation in a dataset) or examining interaction instances (e.g., how much a decision maker relies on an AI assistant) in silos without examining their connections \cite{Wong2023-nx,Vera_Liao2023-gv}. As a result, these approaches do not support the kind of integrated, end-to-end reasoning about hazards that is central to system safety analysis.


 
\subsection{Safety Engineering and System Safety for AI}

Safety engineering frameworks have been used to establish the safety of technical systems as early as the 1940s with the use of \ac{FMEA} and \ac{FTA} in military technology and the Apollo mission \cite{Carlson2012-iq,Swuste2022-xd,Calder1899-gr}. These early frameworks follow an analytically reductive approach where component failures and human errors can be analyzed as a chain of events that lead to an incident. With the increase in complexity of technological systems and specifically the introduction of software, the analytically reductive approach was deemed ineffective in capturing new types of hazards. In the early 2000s, Nancy Leveson proposed an alternative accident model called Systems-Theoretic Accident Model and Processes (STAMP) \cite{Leveson2004-jn}. 
Consistent with system theory \cite{Von_Bertalanffy1972-uw}, STAMP frames safety as an emergent property of a system, resulting from the interactions and control structures among people, technical artifacts/agents, and the environment \cite{Leveson2011-fo}. This paradigm of safety motivated the safety engineering community to look beyond the analysis of units and components, and consider analyzing the interaction between software programs, physical parts, and people needed for establishing a safe system \cite{Leveson2017-di,Leveson1986-ag,Knight2002-dy}. Subsequently, Leveson and Thomas developed \acf{STPA} as a \textit{hazard analysis} method based on STAMP \cite{Leveson_undated-zz}, which was promptly adapted across a number of safety-critical application domains \cite{Park2022-qr,Chen2020-cy,Bas2020-dw}. 

Identifying and mitigating harms from \ac{AI} systems requires a systematic approach that explores and maps out the distribution of existing product development practices beyond an \ac{ML} model and across individuals, automated systems, collectives, and institutional mechanisms \cite{Dobbe2022-ql,Lazar2023-du, Jatho2023-zr, Kroll2021-nk}. Narrowly examining and designing for safety at the \ac{ML} model level is important but not sufficient for ensuring the safety of products that only partially rely on these models \cite{Brundage2020-ck, Khlaaf2023-qs,Mantelero2021-cr}. The system theoretic perspective recognizes that \ac{AI} models do not exist in silos in practical product environments. They are often integrated in a complex network of API calls, other \ac{AI} models, a range of datasets and model outputs, and various user interfaces \cite{Nabavi2023-ce,Widder2022-rf,Crawford2022-qx}. In addition, an \ac{AI} system design is influenced by the decisions of a wide range of actors, including the data curation team, model development team, responsible AI analysts, users, company executives, product managers, and regulators \cite{Sloane2022-ag,Figueras2022-dm,Mantymaki2022-im}. 

Traditionally, \ac{STPA} and system safety have been applied to technologies where there is consensus on what is safe or unsafe (i.e., an airplane that could crash is viewed as unsafe). In contrast, as discussed above, harms arising from \ac{AI} systems are diverse, contested, and often context-dependent, with no single agreed-upon notion of acceptable or unacceptable outcomes. Accordingly, we treat potential harms from \ac{AI} as emergent properties of the system that arise through interactions among technical components, human actors, and institutional processes across development and operation.
Furthermore, \ac{AI} systems have unique differentiating characteristics compared to other complex cyber-physical systems: they can have varying levels of outcome complexity, and the system capabilities \cite{Yang2020-ha} can be probabilistic and distributed. Importantly, these properties are not shared uniformly by all \ac{AI}: they vary markedly across \ac{AI} paradigms---for instance, between an interpretable linear model, an adaptive reinforcement learning agent, and an opaque generative model---so that the same analytical step can demand different considerations depending on the type of \ac{AI} under study. These properties complicate hazard identification and prediction, and must be explicitly considered when applying system-theoretic hazard analysis. These considerations motivate the need to adapt \ac{STPA} for \ac{AI} systems in a way that supports system-level reasoning about harms across the AI development and deployment processes.

\subsection{\acf{STPA}}
\label{STPA_steps}
Below, we provide a brief description of the key steps for conducting \ac{STPA} outlined in the \ac{STPA} handbook \cite{Leveson_undated-zz}. These steps are meant to be performed iteratively and cyclically across the lifecycle of technology development.
 \subsubsection{Identify purpose of the analysis.}\label{identifypurpose} The \ac{STPA} process starts by identifying losses, system boundaries, and hazards. By definition, a \textbf{\textit{loss}} occurs when a stakeholder's value is disregarded or violated. \textit{Hazards} are a \textit{system state or conditions that could be a potential source of harm}. Hazards are defined per \textbf{\textit{system boundary}}, which outlines parts of the system where designers, developers, engineers, or automated agents could exercise some level of control. 
 To provide a simplistic example, a construction company repairing a roof of a house may point to safety considerations (loss of human life) to be important for the project; identify the possibility of workers falling from the steep and bare edges of the rooftop as a fall hazard; and is only able to install safety equipment such as fences within the boundaries of the roof being renovated and not that of the neighbouring house (system boundary).
 
\subsubsection{Create control diagrams.}\label{createctrldiagram} The next step involves creating control diagrams to visually capture the model of the system within its system boundary. It represents the elements and interactions between different elements within the system in the form of a feedback loop called \textit{control feedback loop}. 
The two key elements of a control feedback loop are the \textit{\textbf{controller}} -- an algorithm, a person, or an organization that can make decisions or perform actions to control something -- and the \textit{\textbf{controlled process}} -- the component, process, or another controller being controlled by the controller.
A controller is represented with a \textit{control algorithm} -- a description of how the controller makes decisions -- and \textit{process model} -- the controller's internal/mental model the environment used to make decisions.

The relationship between controllers and controlled processes are represented with two types of interactions: \textit{\textbf{control action}}, which is a constraint that the controller puts on the controlled process, and \textit{\textbf{feedback},} which is the response/feedback the controlled process provides to the controller. 
Typically, the various controller, controlled processes, and their interactions are arranged visually in a hierarchical manner, such that those with the highest level of control/authority appear on top.

\subsubsection{Identify unsafe control actions.} \label{identifyuca} Once a control diagram is drawn, the next step is to identify actions that could be taken by the controller that lead to any of the hazards identified earlier.
These are called \textit{\textbf{\ac{UCA}}}. An \ac{UCA} is fully described using five parts: the controller (\textit{source}) that provides the problematic action (\textit{control action}), the \textit{type} of action/inaction that could lead to hazards (e.g., is it the execution, lack of execution, untimely execution of the control action that can lead to the hazard?), specific system condition or state context (\textit{context}), and \textit{link to hazard}.
\subsubsection{Identify loss scenarios.} \label{identifylossscenario} 
Finally, the circumstances that could lead to a \ac{UCA} -- \textbf{\textit{loss scenarios}} -- are identified. To identify loss scenarios, the analysts need to consider all potential issues with the elements and interactions, such as missing feedback loops, incorrect mental or process models, and improper control algorithms. At the end of this process, the analysts develop a list of safety requirements based on the loss scenarios to provide the necessary constraints and to minimize or eliminate cases of inadequate control.

\section{Methodology}
\label{methodology}
\begin{figure}[t]
  \centering
  \includegraphics[width=0.85\linewidth]{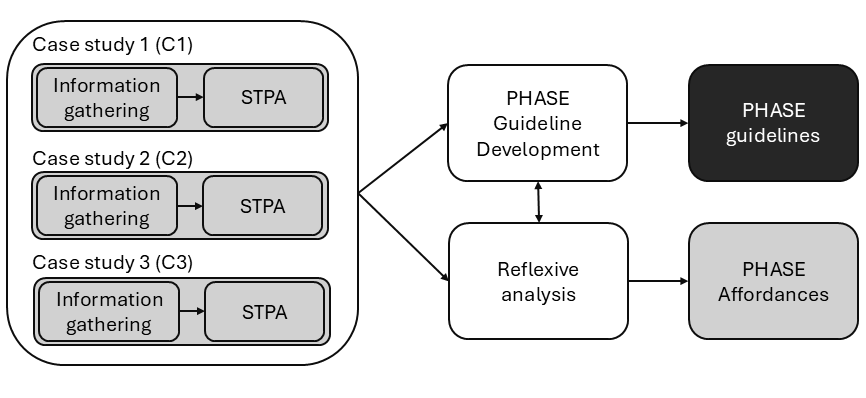}

  \caption{Methodology overview: \ac{STPA} was conducted for each case study after we gathered the necessary information. The information gathering process was different for each case study. Iterative reflexive analysis led to the generation of the \ac{PHASE} guideline and identification of affordances.}
  \label{fig:method}
\end{figure}

Figure \ref{fig:method} presents an overview of our research method. 
As Njie and Asimirian outline, case studies as a research method enable in-depth and first-hand investigation of events and processes \cite{Njie2014-ae}. Therefore, we used case studies to investigate how \ac{STPA} needs to be interpreted when it is used to analyze \ac{AI} systems. 
Through an iterative, reflexive analysis of the case study applications, we developed \ac{PHASE}, a guideline for conducting \ac{STPA} on \ac{AI} systems, and identified the key affordances it provides. We further refined the guidelines’ language based on feedback from a pilot study, described in the supplementary material.
Here, we describe the case studies, the details of applying the \ac{STPA} framework, and the process of developing reflexively analyzing \ac{PHASE}.


\subsection{Case studies}

To illustrate the applicability of \ac{STPA} we chose the following real-world case studies based on the diverse range of \ac{ML} model they represent, application context, and the nature of interaction between the \ac{AI} system and a human user. Specifically, the three cases span distinct \ac{ML} paradigms---supervised linear modelling, reinforcement learning, and transformer-based generative modelling---and distinct interaction modes---human-in-the-loop, human-out-of-the-loop, and human-as-tool-user---so that we could examine whether and how different types of \ac{AI} place different demands on the analysis rather than treating \ac{AI} as a single, homogeneous technology. An additional selection criterion was our ability to work closely with the teams developing these systems, enabling access to design decisions and development processes that are essential for conducting such analysis.
\begin{itemize}
    \item \textbf{Case 1 (C1) - Early warning system:} 
    A linear regression algorithm is being designed in the Netherlands to predict the likelihood of late-onset sepsis (LOS) in pre-term infants before a clinical suspicion \cite{Van_den_Berg2023-oy}. It is to be used as a proactive warning system alerting clinicians to start early diagnostic testing for LOS and, ultimately, provide timely treatment. 
    As a human-in-the-loop system, a clinician makes final, clinical decisions.
    \item \textbf{Case 2 (C2) - Insulin Injection:} An \ac{RL}-based, automated and personalized insulin injection system is being developed in Canada for patients with diabetes \cite{Basu2023-er}. 
    Some diabetic patients use an artificial pancreas that facilitates automatic insulin injection. 
    The \ac{RL} algorithm is being developed for use in the artificial pancreas to allow tailored insulin doses for each individual in a human-out-of-the-loop manner.

    \item \textbf{Case 3 (C3) - Storyboarding :} Some \ac{ML} artists have adapted \ac{T2I} models as part of their creative practice, such as storyboarding for video creation \cite{Yu2022-ed, Franceschelli2021-kj,Saharia2022-em, Li2022-wj}. In this case study, we focus on \ac{T2I} demo platforms generally. While we do not focus on a specific platform, many such platforms are produced by companies in the United States, and rely on some combination of transformer-based or diffusion architecture \ac{T2I} generative models.

\end{itemize}

\subsection{Applying the \ac{STPA} framework}
To apply \ac{STPA}, we started by gathering information about the \ac{AI} system, its stakeholders, and how the system was developed and integrated in the given context. To do so, we connected with the relevant stakeholders of each of the projects and conducted secondary research. 
Considering time limitations, data availability, and access to subject-matter experts, the information-gathering process was customized for each case. For C1, the lead author held a preliminary meeting with the \ac{ML} developer and data scientist to understand the project's objectives and used a peer-reviewed academic publication on the project as the main reference for the analysis. For C2, the lead author held monthly meetings for eight months with two \ac{RL} researchers -- one of whom developed \ac{RL} algorithms for insulin injection.
For C3, the lead author conducted eight interviews and one workshop (15 participants) involving experts along the \ac{T2I} model development process and \ac{ML} artists who use these models in their creative practice. 

After information gathering, the lead author followed the \ac{STPA} steps outlined earlier (Section \ref{STPA_steps}) and applied them for each case study. While the procedural steps are those of standard \ac{STPA}, the interpretive work at each step---deciding what counts as a loss, where to draw system boundaries, which components act as controllers, and what makes a control action unsafe---is where \ac{AI}-specific considerations entered and where the case studies differed from one another; Section~\ref{interpretations} details these adaptations. The lead author then cross-checked the hazards identified in the analysis against relevant literature and subject-matter experts to sanity-check the results from the \ac{STPA}. The co-authors of the paper also reviewed each of the analyses and provided feedback on the application of \ac{STPA}. 

\subsection{Developing the \ac{PHASE} guideline and examining its affordances}

Through reflexive analysis of the case studies and the process of applying \ac{STPA}, the lead author developed \acf{PHASE}, a step-by-step guideline for conducting system-level hazard analysis of algorithmic harms in \ac{AI} systems. This work was informed by iterative discussions and feedback with the co-authors, and the joint reflexive examination of both the case study findings and the analysis process enabled the identification of the key affordances that \ac{PHASE} provides for analysts. The guideline was further refined based on feedback from a pilot study, described in the supplementary material.

%

\section{PHASE Guideline: Interpreting STPA for AI System Case Studies}
\label{interpretations}
Next, we present observations from applying \ac{STPA} across the three case studies and highlight features unique to \ac{PHASE}. Before turning to the individual steps, we briefly summarize how \ac{PHASE} adapts each of the four \ac{STPA} steps for \ac{AI} systems.

In Step 1 (identify the purpose of the analysis), \ac{PHASE} broadens how losses are conceived by drawing on value-sensitive design and by surfacing the value tensions among stakeholders, so that sociotechnical losses are considered alongside safety-critical and performance-related ones; it also draws system boundaries around the development and use processes of \ac{ML}-based systems rather than around the model itself, since developers exercise control over data and design choices but not over what a model learns. In Step 2 (create a control diagram), \ac{PHASE} accommodates multiple and dynamic control structures, because the way \ac{ML}-based systems are used is fluid---a component can act as a controlled process in one context and as a controller in another---and it introduces components specific to \ac{AI} systems, such as datasets, models, safety filters, and learning agents, as controllers or controlled processes. In Step 3 (identify unsafe control actions), \ac{PHASE} summarizes the unsafe control actions characteristic of \ac{AI} systems---functional failures, poor design decisions or misuse, and coordination or communication breakdowns; what distinguishes these from unsafe control actions in traditional \ac{STPA} is that their unsafeness frequently depends on sociotechnical context rather than on the technical timing or sequencing of a control action. Finally, in Step 4 (identify loss scenarios), \ac{PHASE} captures the loss scenarios also found in conventional \ac{STPA}---missing feedback loops, incorrect mental or process models, and technical malfunctions---but adds a class unique to \ac{AI}: loss scenarios arising from emerging and evolving system capabilities, where changes such as retraining, updates, or expanded use outpace existing safeguards, a source of hazard that traditional \ac{STPA}, applied to systems with fixed capabilities, does not typically consider.

The \ac{PHASE} guideline, along with a detailed summary of the case studies, is provided in the supplemental material. 

\subsection{Step 1: Identify purpose of analysis}
\label{ML_identifypurpose}

\begin{table*}[]
\caption{The losses identified across the three case studies. The three types of losses (safety-critical, performance-related, and sociotechnical harms) are identified with \#.}
\label{tab:loss}
\resizebox{\linewidth}{!}{%
\begin{tabular}{ll}
\hline
Case 1 (C1) -Early alert system & \begin{tabular}[c]{@{}l@{}}L1. Loss of life or injury to the preterm infants  \textcolor{red}{\#safety-critical}\\ L2. Loss of time (in terms of efficiency) \textcolor{blue}{ \#performance-related}\\ L3. Loss of reputation/credibility for the physician in charge of a case \textcolor{blue}{ \#performance-related} \\ L4. Loss of patient privacy (regulation about medical data) \textcolor{purple}{\#sociotechnicalharm}\end{tabular} \\ \hline
Case 2 (C2) - Insulin injection & \begin{tabular}[c]{@{}l@{}}L1. Loss of life or injury to the diabetic patient \textcolor{red}{\#safetycritical}\\ L2. Loss of quality of life (i.e., comfort and autonomy) \textcolor{purple}{\#sociotechnicalharm}\\ L3. Loss of efficiency \textcolor{blue}{ \#performance-related}\\ L4. Loss of money \textcolor{blue}{ \#performance-related}\end{tabular} \\ \hline
Case 3 (C3) - Story boarding & \begin{tabular}[c]{@{}l@{}}L1. Loss of creativity \textcolor{purple}{\#sociotechnicalharm}\\ L2. Loss of diversity \textcolor{purple}{\#sociotechnicalharm}\\ L3. Loss of accessibility \textcolor{purple}{\#sociotechnicalharm}\\ L4. Loss of efficiency \textcolor{blue}{ \#performance-related}\\ L5. Loss of quality \textcolor{blue}{ \#performance-related}\\ L6. Loss of reputation \textcolor{blue}{ \#performance-related}\end{tabular} \\ \hline
\end{tabular}%
}
\end{table*}

\begin{figure*}[]
  \centering
 \includegraphics[width=0.8\textwidth]{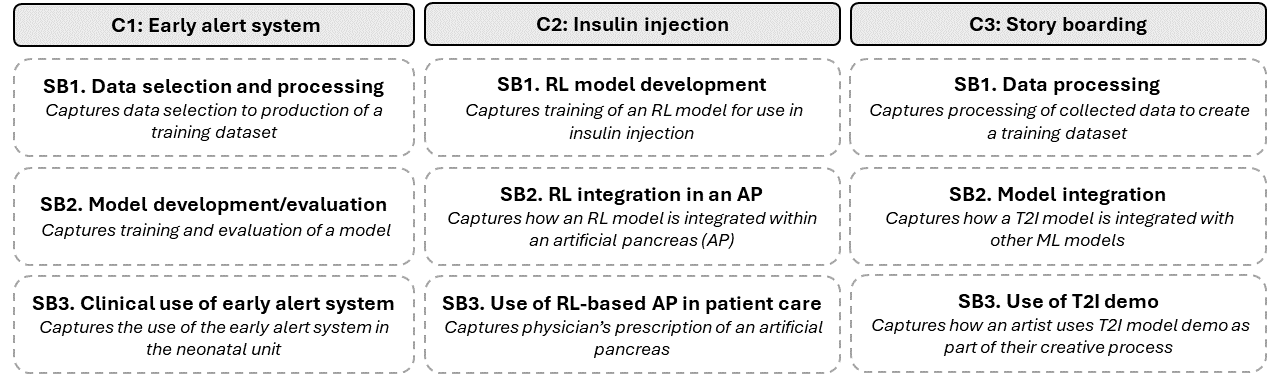}
  \caption{Three \textit{system boundaries} (SB) defined for each of the case studies (C). In practice, analysts are free to define SBs following Section \ref{identifypurpose}. The items under the Case title describe the scope of the SB we defined for the Case Study.}
  \label{fig:systemboundary}
\end{figure*}

Applying the first step of \ac{STPA} - \textit{identify purpose of analysis} - to \ac{AI} systems required a substantive reinterpretation. In traditional applications of \ac{STPA}, this step typically centers on defining safety-critical losses and establishing system boundaries around controllable technical components. Our case studies showed that, for \ac{AI} systems, this framing is too narrow: both the notion of loss and the selection of system boundaries are deeply shaped by sociotechnical context, value tensions, and the processes through which AI systems are developed and operated. In response, \ac{PHASE} adapts the first step of \ac{STPA} by broadening the interpretation of loss to incorporate insights from value-sensitive design and by emphasizing process-oriented system boundaries aligned with the algorithmic supply chain. The sections below describe our observations from the case studies and how they informed these adaptations. 

\paragraph{Identifying losses}
The losses identified for each case study go beyond the traditional definition of safety (e.g., loss of life, damage to property or environment). As shown in Table \ref{tab:loss}, the identified losses consider a broader set of emergent harms from \ac{AI} systems. Our reflexive analysis reveals that there are three different categories of losses pertinent in an \ac{STPA} of an \ac{ML} system: 1) traditional safety-critical losses, 2) performance-related losses, 3) sociotechnical losses. \textit{Safety-critical losses} represent losses in the traditional definition of safety, such as loss of human life, property damage, and the environment (\textit{C1.L1}). \textit{Performance-related losses}, on the other hand, emerge when the \ac{AI} system does not meet performance expectations (\textit{C1.L2}) or when the individuals, groups, or organizations experience loss of reputation or economic loss (\textit{C3.L6},\textit{ C2.L4}). Lastly, \textit{sociotechnical losses} cover a broad range of sociotechnical harms that individuals and groups can experience that may not be safety-critical or performance-related (\textit{C1.L4}, \textit{C2.L2}, \textit{C3.L2}). Considering the growing body of research in algorithmic harms and our findings from the case studies, we found it much more appropriate to use existing taxonomies \cite{Weidinger2022-ni,Shelby2023-to,bird2023}, industry-specific regulatory requirements \cite{euAIact}, and community-based and participatory research methods \cite{DeVos2022-jk,Wang2022-cx} for loss identification of \ac{AI} systems. 
\paragraph{Defining system boundaries} 
Our selection of system boundaries was guided by the structure of the algorithmic supply chain and by the information available about each case, as well as by where stakeholders indicated that they could meaningfully exercise control. Rather than centering boundaries on individual models, we drew them around development and use processes, which enabled a more thorough analysis of how hazards emerge across the lifecycle of an \ac{AI} system. The number of system boundaries is not prescribed by \ac{PHASE}; analysts define as many as are needed to cover the loci of control relevant to the identified losses. In our cases, three boundaries per system happened to capture the accessible points of control, but this is illustrative rather than a requirement, and prior work has shown that the \ac{ML} development lifecycle can be decomposed into many more stages. The boundaries also differ across cases---for example, integration or a distinct operational boundary appears in some systems but not others---because the systems, their paradigms, and the points at which developers, users, or automated agents reported having decision-making authority differ. Importantly, drawing boundaries around sub-processes does not discard the whole-system view: because boundaries are connected through shared control actions and feedback, interactions across them remain visible rather than being lost, and a hazard localized in one boundary can propagate through these links to losses defined in another.

Across all cases, we identified system boundaries within data collection and processing (e.g., \textit{C1.SB1}, \textit{C3.SB1}), where developers made decisions about data sources, feature inclusion, and labeling; and within model development, evaluation, and integration (e.g., \textit{C1.SB2},\textit{ C2.SB1},\textit{ C3.SB2}), where choices about learning algorithms, architectures, metrics, and outputs were exercised. However, despite this apparent control, stakeholders consistently noted limited visibility into—and control over—the learned latent representations of \ac{ML} models, which constrained the usefulness of drawing system boundaries narrowly around the model itself.

The contrast between the reinforcement learning (RL) case and systems built around opaque, black-box generative models further illustrates this point. In the RL case, we identified an operational system boundary (\textit{C2.SB2}) where control shifted from human operators to the algorithmic agent, which autonomously adjusted the amount of insulin delivered to a diabetic patient. This clear locus of control during operation made it possible to reason explicitly about hazards arising from the system’s autonomous decision-making. In contrast, for opaque generative systems such as text-to-image models, stakeholders reported limited ability to interpret or directly control model behavior at inference time. For a general-purpose system such as a \ac{T2I} platform, the space of possible uses is open-ended, which makes boundary definition especially consequential: a boundary drawn loosely around ``use of the demo'' is too vague to determine whether a given hazard is controllable within it. We therefore drew the downstream boundary narrowly around the generation and dissemination of a specific output (e.g., a storyboard used for video creation, \textit{C3.SB3}), where a hazard such as the dissemination of false or harmful content becomes controllable---through the prompts entered, the safety filters applied, and the decision to publish---rather than around the platform's use in the abstract. Consequently, system boundaries for these systems were more meaningfully drawn around upstream development processes and downstream use contexts—such as content generation and dissemination (e.g., \textit{C3.SB3}) - where human actors retained control over how outputs were produced, interpreted, and shared.

\paragraph{Outlining Hazards}
Hazard identification largely followed established \ac{STPA} practice and was driven by the system boundaries and loss definitions established in the first step of the analysis. The hazards identified across the case studies aligned with the three loss categories described above and often spanned multiple losses. For example, the hazard “The system creates a model or output that does not meet performance requirements” for \textit{C1.SB2} could degrade an early-warning system and, in extreme cases, contribute to loss of life (\textit{C1.L1}). Other hazards reflected well-documented sociotechnical concerns in the \ac{FATE} and RAI literature, such as lack of interpretability (\textit{C1.SB2}; \textit{L1}, \textit{L3}) \cite{Bhatt_Surabhi_undated-ye} and the dissemination of false or harmful content (\textit{C3.SB3}; \textit{L5}, \textit{L6}) \cite{Birhane2022-vp}. Here, the input and output safety filters referenced throughout the \ac{T2I} case are classifiers that screen prompts and generated images to block disallowed content; their performance refers to how accurately they do so, that is, their false-positive rate (blocking benign content) and false-negative rate (allowing harmful content through), and a poorly designed filter is one whose threshold admits too many harmful outputs or suppresses too much legitimate use.

\subsection{Step 2: Create a control diagram }
\label{ML_controldiagram}
A key challenge in applying \ac{STPA} to \ac{AI} systems is that some components do not occupy fixed roles as controllers or controlled processes. For example, an AI assistant may function as a controlled process in one context (e.g., supporting human decision-making) and effectively act as a controller in another when users are organizationally required to follow its outputs. In \ac{PHASE}, we explicitly account for this role fluidity by encouraging analysts to model control relationships as context-dependent rather than static. Capturing this role fluidity at the control-diagram stage is essential, as hazards depend not only on system behavior but also on how control shifts between humans and \ac{AI} systems in practice.

\begin{figure*}[!tbp]
  \centering
  \includegraphics[width=0.7\linewidth]{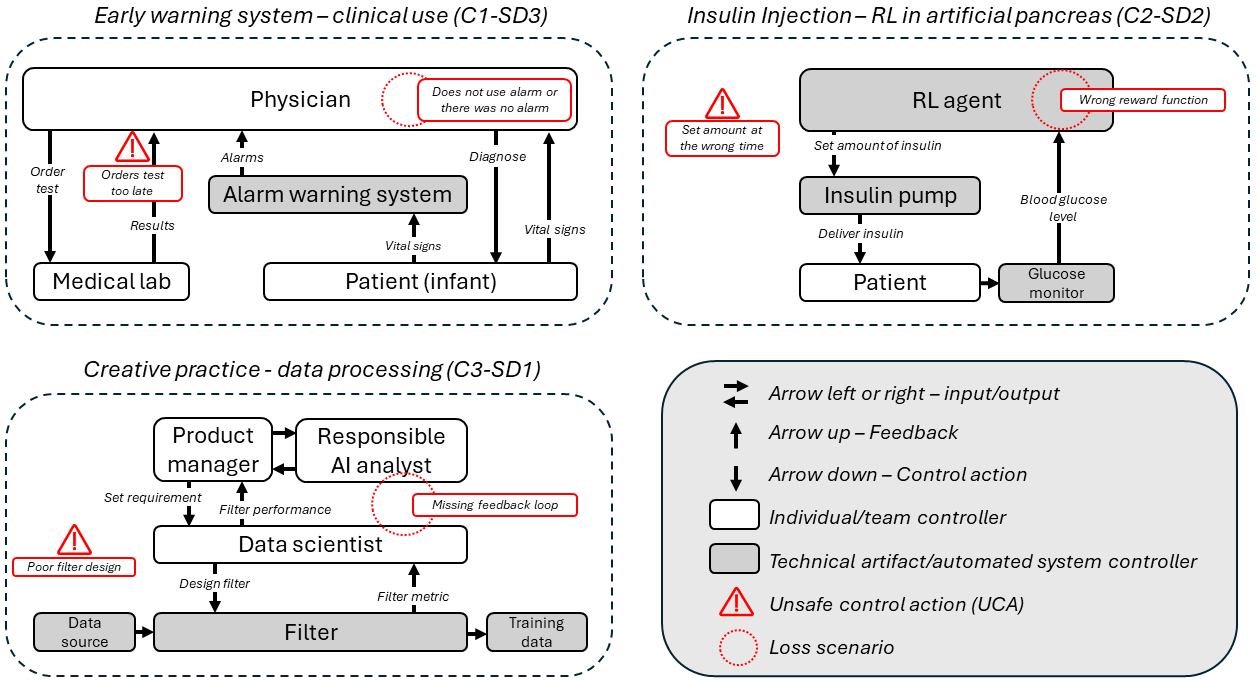}
  \caption{This diagram illustrates the simplified control structure. Each of the System Diagrams (SD) corresponds to one System Boundary (SB). We have included only three of the system boundaries in this diagram,  one from each case study.}
  \label{fig:controldiagram}
\end{figure*}

Using the core concepts of \ac{STPA}, we mapped controllers, controlled processes, control actions, feedback, inputs, outputs, and control hierarchies within each selected system boundary. Across the case studies, we were able to translate these concepts to \ac{AI} systems and interpret how they manifest across both development and operational contexts. We constructed a control diagram for each system boundary, yielding a total of 9 diagrams; the full set is provided in the supplementary material, and we show one boundary per case study in Figure \ref{fig:controldiagram} to illustrate the range of control structures across the three paradigms. To read such a diagram, one traces each control action downward---the constraints a controller imposes on the process below it---and each feedback signal upward---the information returned to the controller---so that, together, they reveal where control is adequate and where it may break down. Figure \ref{fig:controldiagram} presents one representative control diagram from each case study. The elements represented as controllers or controlled processes ranged from individuals and teams to technical artifacts. In many cases, controllers were individuals or groups responsible for decision-making during development or use. For example, in \textit{C1.SB1}, controllers included the data collection team, data processing team, and hospital database manager, while the collected dataset functioned as a controlled process. Similarly, in \textit{C3.SB2}, technical artifacts such as input and output safety filters and text-to-image models were modeled as controlled processes. However, individuals and teams were not always the primary controllers. In some contexts, control was exercised by technical artifacts themselves. For instance, in \textit{C2.SB3}, the reinforcement learning agent and the insulin pump acted as controllers by autonomously determining insulin delivery. In other cases, control relationships depended on organizational and use-context constraints. For example, in \textit{C1.SB3}, when use of an early alert system was mandated for physicians, the \ac{AI} system constrained clinical decision-making, effectively positioning it as a controller despite its nominal role as a decision-support tool. These examples illustrate why treating \ac{AI} systems as having fixed roles obscures important sources of risk.

The top-to-bottom representation of the control hierarchy (with higher levels indicating greater control) was generally effective for visualizing these relationships, but the ordering of the hierarchy was often context-dependent and subject to change based on system design choices, organizational policies, or patterns of use. Across the case studies, control actions ranged from development-time decisions (e.g., dataset processing and model design choices), to use-time actions (e.g., prompts entered by ML artists interacting with a text-to-image system), to technical constraints (e.g., threshold settings for safety filters). Feedback took the form of evaluation metrics, sensor outputs, user feedback, or other forms of human communication. Inputs and outputs captured both technical artifacts (e.g., training datasets) and coordination among individuals and teams, such as collaboration between physicians and developers to ensure clinically appropriate model behavior (\textit{C1}). 

\subsection{Step 3: Identify unsafe control actions }
\label{ML_uca}
Overall, we found that Step 3 of \ac{STPA}—identifying unsafe control actions (\acp{UCA})—could be applied to \ac{AI} systems with minimal modification. The core \ac{STPA} logic of examining control actions in relation to system-level hazards was transferred directly; however, careful interpretation of the four conditions under which a control action becomes unsafe was essential in AI contexts. In particular, whether a control action is provided, not provided, provided too early or too late, or applied too long or too briefly often depended on sociotechnical context rather than purely technical timing or sequencing. From our analysis, we identified three main trends in \ac{UCA}s applicable to AI systems. 

\textbf{Functional \acp{UCA}}
The most common type involved \textit{functional \acp{UCA}}, where inadequate control resulted from a malfunction or dysfunction of the controller. For example, in \textit{C2.SB2}, the insulin pump in the artificial pancreas failed to release the required amount of insulin at a given time. Similarly, in \textit{C1.SB3}, an incorrect alarm from the early alert system could negatively influence a physician’s decision-making. These \acp{UCA} were primarily associated with performance-related and safety-critical losses.

\textbf{\acp{UCA} from poor design decision-making or misuse}
A second type of \acp{UCA} stemmed from \textit{poor design decisions or misuse} by experts, system developers, or users. For instance, in \textit{C3.SB2}, model developers could select an inappropriate safety filter threshold, enabling harmful prompts or image outputs. In \textit{C1.SB1}, the data processing team could select incorrect clinical variables as features, leading to misleading model behavior. These \acp{UCA} highlight how design-time decisions can render otherwise correct control actions unsafe in context. 

\textbf{\acp{UCA} from communication/coordination} The final type of \acp{UCA} involved \textit{inadequate organizational coordination or miscommunication}. For example, model developers might receive incomplete or incorrect safety requirements, or lack sufficient time to address them before deployment deadlines (\textit{C3.SB2}). In these cases, unsafe control actions emerged not from technical failure, but from misaligned processes and constraints across teams.

\subsection{Step 4: Identify loss scenarios}
\label{ML_losscenario}
 
The original \ac{STPA} process was highly effective in the process of understanding loss scenarios for \ac{AI} systems. We brainstormed a list of loss scenarios by thinking through all the elements of the control diagram. A diverse range of loss scenarios were identified, which included organizational/institutional causal scenarios (i.e., poor understanding or communication), interaction issues (e.g., incorrect mental model) and technical challenges (e.g., missing metrics/evaluations of a system). Across the case studies, however, we observed a recurring class of loss scenarios unique to \ac{AI} systems: losses arising from emerging and evolving system capabilities. These scenarios did not necessarily stem from immediate component failure, but from changes in model behavior over time—such as updates, retraining, or expanded use—that outpaced existing assumptions, safeguards, or organizational processes.  
In addition, we observed two other prevalent forms of loss scenarios. A common institutional causal scenario across our case studies involved missing feedback loops between different teams that are in charge of various parts of the development process. For instance, often model development teams use pre-existing datasets that have been collected and processed by another group and at a different time frame than when the model (e.g., \ac{T2I} model) is being developed. In such cases, there is a missing feedback loop between the developers of the model and creators of the dataset (e.g., \textit{C3-SB2}). Another type of causal scenario is when the user has an incorrect mental model of the \ac{AI} system, meaning that they do not understand how to use it correctly or how it works. This scenario is characteristic of \ac{AI} systems deployed to non-expert users: unlike traditional safety-critical software operated by trained specialists, \ac{AI} tools are increasingly used by people without an accurate model of the system's competence or limits, and \ac{PHASE} captures this through a mismatch between the controller's process model and the actual behavior of the controlled process. For instance, for \textit{C1.SB3}, if the physician does not understand the reason for the alarm, they could make the wrong lab order or diagnosis. The third type of loss scenario is \textit{technical challenges} which occurs when a component (i.e., a pump, \ac{ML} model, safety classifier) is functioning erroneously.
This could lead to inadequate control when making decisions about a patient's health. The type of loss scenarios we identified is not exhaustive. However, they provide an understanding of the type of causal loss scenarios for inadequate control in \ac{AI} systems.

\section{Affordances of System Theoretic Perspective for algorithmic harm from AI systems}
\label{affordances}
Based on our reflexive analysis, we identified four affordances observed across our case studies that build on \ac{STPA} when applied to \ac{AI} systems using our \ac{PHASE} guideline. We stress that the value here is not system-level analysis in itself---which \ac{STPA} already provides---but the specific adaptations \ac{PHASE} makes for \ac{AI}: an expanded loss taxonomy that includes sociotechnical harms, the treatment of \ac{AI} components as context-dependent controllers or controlled processes rather than fixed-role artifacts, and loss scenarios that account for evolving model capabilities. The affordances described below follow from these adaptations, and it is these adaptations, rather than the general system-centric stance, that we claim as \ac{PHASE}'s contribution.

\subsection{Affordance 1: Analyze AI at the systems level}
\label{aff1}
Since \ac{STPA} is designed to analyze a process or an artifact at the systems level, \ac{PHASE} inherits this affordance. Prior work suggests that algorithmic harms from \ac{AI} arise from system-wide issues \cite{Smuha2021-cp,Selbst2021-pk,Thomasen2023-cc}. For instance, representational harms are system-level issues that ``occur when algorithmic systems reinforce the subordination of social groups along the lines of identity, such as disability, gender, race and ethnicity, religion, and sexuality'' \cite{Shelby2023-to,Karizat2021-ch}. They arise from a combination of ad-hoc factors such as inaccurate \ac{ML} model performance for minority groups, misuse of the system in a given application, or exclusionary social norms. 
Instead of narrowly focusing on ``fixing" an unbalanced training dataset, system-level analysis prompts us to examine how this dataset will interact with other parts of the system, such dataset analysts, other \ac{ML} models, or the user interface \cite{Barocas2021-sr,Wang2022-ck}. 

\ac{STPA} is a systematic method that establishes possible causal chains between undesirable outcomes of the system and inadequate control. As such, the \textit{losses} and \textit{harms} identified in the first step of \ac{PHASE} (Section \ref{ML_identifypurpose}) define the negative impact of the system an analyst aims to prevent or mitigate. Subsequently, \acp{UCA} (Section \ref{ML_uca}) explicitly defines all possible cases of inadequate control in the chosen system boundary that lead to a hazard. Loss scenarios (Section \ref{ML_losscenario}) finally establish the causal link connecting the undesirable outcomes to specific unsafe states of the system. For example, we identified the hazard \textit{``H: Clinician misdiagnoses the patient."} to stem from \textit{inadequate control} over the proper use of the early alarm system and timing for ordering a blood test, which can lead to \textit{L: loss of life}. 
In this case, the control action of \textit{``clinical use of the early alert system''} could be unsafe and in several loss scenarios, such as when the clinician does not understand how to use the early alert system. Notably, this hazard would not surface from a component-level analysis of the model or dataset alone: a fairness or accuracy audit of the alert model in isolation could report acceptable performance, yet the loss still arises from the interaction between an imperfect alert and a clinician who over-trusts or misreads it. It is only by analyzing the dataset, model, and user together that this system-level hazard becomes visible. 


\subsection{Affordance 2: Accounting for social factors and sociotechnical harms}
\label{aff2}
Formalizing losses at the beginning of \ac{STPA} mobilizes a discussion and instantiates the need for consensus amongst stakeholders about harms that must be prioritized.
Amongst practitioners and scholars, there is a divergence of perspectives on the types of harm that need to be assessed and mitigated for \ac{AI} systems \cite{Saetra2023-jq,Lazar2023-du, Hanna_undated-wm}. Currently, company policies and ad-hoc conversations amongst limited, invited individuals shape practitioners' choices on what harms need to be minimized or fully prevented \cite{openAIsafetypolicy}. However, following \ac{PHASE} structures the analysis so that a broad set of sociotechnical harms is considered at the outset. It cannot compel an analyst to be exhaustive; rather, by making the enumeration of losses the explicit first step, it renders the omission of such harms visible and open to challenge rather than leaving it implicit. Furthermore, in defining the system boundary, \ac{PHASE} connects the losses to a network of individuals, teams, technical artifacts, automated systems, and institutional mechanisms which form a system boundary along the \ac{AI} development cycle as opposed to just examining a loss by focusing on one of these elements (i.e. blaming a developer or assuming \ac{ML} models cannot be explained).  
Lastly, the control diagram and its components in \ac{PHASE} provide the conceptual flexibility to map social and technical elements within a system boundary. For example, a controller could be a human (e.g., an artist) who has a specific mental model of how to use a \ac{T2I} platform. It could also be an \ac{RL} agent with a map of the environment (\textit{process model}) and a reward model (\textit{control algorithm}) that guides the suggested dosage of insulin (\textit{control actions}). This conceptual flexibility allows for the integration and prioritization of sociotechnical factors in \ac{AI} harms analysis.

\subsection{Affordance 3: Identify accountability chain by establishing traceability}
\label{aff3}
Traceability in harm reduction is a pressing concern in the AI community. Recent studies indicate that responsible AI practitioners are frustrated by the lack of systematic methods that establish a connection between a specific set of harms and the means to mitigate the harm \cite{Madaio2022-vt,Rakova2021-dg,Sloane2022-ag}. This problem is exacerbated by the fact that many systems are developed by a multitude of individuals and groups across institutions that hardly interact directly with one another \cite{Widder2022-rf,Nabavi2023-ce}. That is, many unsafe system states can accumulate between individual decision-makers, the technical system, and the development process without the situation being detected or managed \cite{Widder2022-rf,Kroll2021-nk}. As described earlier, the \ac{STPA} framework allows us to connect harms to inadequate control within a system. Furthermore, it allows an analyst to make explicit who has control over what, and make the harms and hazardous states of a system traceable for everyone. Specifically, the concept of hierarchical control structure as interpreted in \ac{PHASE} engages the analyst to map who or what has control in relation to other actors or artifacts by mapping their interaction using control action and feedback. 
For instance, by drawing the control structure, we needed to examine whether the clinician had control over the early alert system meaning that they could choose to use the warning upon reception (i.e., feedback) or that they needed to use the warning from the early alert system (i.e., the control action) by thinking through how the early warning system is integrated into clinical practice. Understanding the control hierarchy fosters an understanding of how power, control, and responsibility are distributed in a system, which ultimately results in a more managed chain of accountability for responsible AI deployment. Concretely, this is what lets an analyst localize a harm to its source. If the early alert system contributes to a harmful misdiagnosis, tracing the loss scenario back through the control structure distinguishes whether the inadequate control originated in data collection (for example, a context-specific perturbation missing from the training data, surfaced as a missing feedback loop between the data and model teams), in the model (a training deficiency, surfaced as an inadequate control algorithm), or in use (a clinician who does not understand the alert, surfaced as an incorrect process model). Because a single loss scenario can implicate more than one control action or feedback link at once, the same structure also makes explicit when a harm is combinatorial---arising from the joint effect of, say, a data gap and a user's misunderstanding---rather than attributable to any one component in isolation.

\subsection{Affordance 4: Monitor and address emergent hazards over time}

\ac{STPA} framework enables an analyst to monitor and account for changes in individuals, teams, technical artifacts, and institutional mechanisms over time. By emergent hazards over time, we mean hazards that were absent or not analyzable at initial design but arise as the model, its use, or its organizational context changes---for example, as a reinforcement learning agent adapts its policy during operation or as a generative model is retrained or updated after deployment. As C2 illustrates, this includes the changes that stem from the adaptivity of \ac{AI} systems such as \ac{RL}. These adaptations can be reflected in various components of a control diagram, such as control actions, feedback loops, and elements of a controller (e.g., decision-making protocol, process/mental model of the world). For instance, the change in the output of an \ac{AI} model could be captured in a control action as constraints/information a clinician or an artist receives. Furthermore, changes in an \ac{AI} model's latent representation could be captured in the process model of the controller or controlled process (depending on whether a model is acting as a controller or is a controlled process). Changes in how individuals, teams, and organizations perceive and operate, such as changes in how an \ac{AI} model is evaluated, could be captured as part of the feedback loop between the model and the developer team. Such concrete means of accounting for changes in the system allow analysts to monitor the dynamics systematically and iteratively, prompting them to examine emergent hazards as \ac{AI} systems change in how they are integrated or interact with their sociotechnical context. This is a critical affordance given the rapid pace of \ac{AI} development and increased interest in adaptive and agentic \ac{AI} systems \cite{Shavit_undated-tf, Weidinger2023-pe}. 

\section{Discussion}
\label{discussion}
In this paper, we employed a case study approach to translate \ac{STPA} for the analysis of AI systems.
We illustrated that \ac{STPA} framework can not only be translated to analyze \ac{AI} systems and identify potential sources of algorithmic harm, but that the \ac{PHASE} implementation of the analysis enables four affordances critically missing in existing AI assessment frameworks. 
\ac{STPA} framework allows analysts to map out sociotechnical system dynamics, and instantiate accountability by examining harms at the system level and tracing them to control methods for harm reduction. Moreover, the framework allows for accounting of changes in individuals, collectives, \ac{AI} models and artifacts, and institutional processes that could happen over time. Here, we discuss the broader implications of adopting a system-theoretic perspective for \ac{AI} governance and building a responsible \ac{AI} culture in organizations. We also elaborate on limitations and avenues for future work. 

\subsection{Implications for policy and AI governance}
From self-governance mechanisms in corporations to national regulations on \ac{AI}, there has been a growing effort to practice due diligence and establish accountability mechanisms across the \ac{AI} industry \cite{The_White_House2023-ad,euAIact}.
As illustrated in our findings (Section \ref{aff3}), establishing traceability between harms and causal scenarios can be a powerful tool toward this end. 
It can enable an analyst 
to define a division of accountability where specific roles/groups are held responsible for maintaining adequate/safe control of certain sub-systems \cite{Leveson2011-fo}. 
Although our system boundaries did not explicitly focus on internal or external governance mechanisms, some of the controls added by internal governance mechanisms were observed in our case studies. 
For example, for \textit{C3-SB1} as captured in Figure \ref{fig:controldiagram}, the responsible AI analyst provides input on safety requirements to product managers about datasets, which are then passed onto the data scientists. These safety requirements would typically be developed based on existing regulations or internal company policies \cite{openAIsafetypolicy}. Given that 
\ac{STPA} has already been used in other domains to examine and improve governance mechanisms; we expect \ac{PHASE} to be able to afford the same function for \ac{AI} systems \cite{Leveson2004-jn}. 

For example, such a process could point out the missing feedback loop between the data scientist and the responsible AI analyst in \textit{C3-SB1}, which hints at the lack of quality control of the training dataset and the potential need for a data quality approval mechanism. In addition to improving internal accountability protocols for a company, external actors such as standard organizations could leverage these affordances to establish best practices for conducting hazard analysis, and mobilize policymakers to refer to these standards in the regulation for certain \ac{AI} products \cite{Suo2017-hc}.

\subsection{Building a safety culture: systems lens for responsible and safe AI}

Given the focus of our three case studies in the product development context, we also observed how existing culture within the organization and management practices can contribute to sociotechnical harm. For example, in \textit{C3-SB1}, the loss scenarios for the \ac{UCA} of poor safety filter design could stem from a responsible AI analyst not having enough time to review the dataset and provide safety requirements to the product manager before the product needs to be launched. The presence of toxic or profit-driven culture within the tech industry and its relationship to the quality of products deployed in the market is not new \cite{Moss2020-xl}. \ac{PHASE} helps account for this social factor and its relationship to a specific set of harms. 
Our hope is that such explicit acknowledgment of harms that can stem from social contexts can help organizations prioritize the implementation of a safety culture. Even though \ac{STPA} could be a tool for building a safety culture, adoption of \ac{STPA} demands a company culture that values safety and is willing to allocate resources to it. Lack of incentives and financial/human resources limits the applicability and effectiveness of many of the existing responsible AI/AI ethics frameworks \cite{Madaio2022-vt,Rakova2021-dg}. The persistence of these challenges for \ac{STPA} would hinder adoption even with the development of guidelines that translate and illustrate the framework for RAI practice. 

\subsection{Limitations and future work}

Recent discussion about the possible risks of developing agentic systems suggests that agentic \ac{AI} will be inherently dangerous considering that lack of any meaningful methods to control these systems and comprehensively understand their process model of the world \cite{Russell2022-hc, Shavit_undated-tf, Chan2023-rk}. While none of our case studies involve an agentic \ac{AI}, it was clear that the \ac{PHASE} effectively maps where and what humans and technical systems/processes \textit{have or do not have} control. We believe that this mapping can offer valuable insights in contextually examining increasingly agentic systems, especially in investigating areas of inadequate or lack of control to examine systems that are inherently unsafe or uncontrollable.

There are many known limitations of \ac{STPA} that also transfer to \ac{PHASE}. First, \ac{STPA} frames interaction between processes or agents as control and feedback. Such framing limits how nuances in human relationships can be captured, such as collaborative decision-making and social influence between agents. Furthermore, similar to any analytical process, analysts must make normative choices, assumptions, and simplifications in conducting \ac{PHASE}.
This means that the views, values, and perspectives of the analysts affect the final analysis. Relatedly, the conceptual flexibility that lets analysts map social and technical elements within a boundary is also a limitation: analysts may equally choose not to include such elements, so the same flexibility that is a strength for diverse use cases can weaken an analysis in less careful hands. To improve robustness against this analyst-dependence, \ac{PHASE} can be conducted by multiple or deliberately diverse analysts and the resulting hazard and loss-scenario lists cross-validated against one another. Therefore, it is imperative to have well-established industry standards on what a quality \ac{PHASE} should look like. We provide a running example in the supplementary material as a starting point. A further limitation concerns generality: our findings rest on three case studies, and although these were chosen to span distinct \ac{ML} paradigms and interaction modes, they cannot represent the full \ac{AI} landscape, so broader application is needed to test how far these adaptations generalize. Lastly, \ac{PHASE} is a comprehensive hazard analysis that requires adequate expertise and time; to lower this barrier, the supplementary guideline provides step-by-step instructions, templates, and a worked example, though effective use still assumes basic familiarity with the system being analyzed and with control-based reasoning. 

\section{Conclusion}
\label{conclusion}
In this work, we interpret and translate \ac{STPA}, a well-established system safety hazard analysis framework.
We find that system safety models and methods offer a complementary perspective to existing responsible \ac{AI} efforts. It helps us to 
defining safety as an emergent property of the system. 
Through three real-life case studies, we conclude that the concepts and steps of \ac{STPA} apply readily to the analysis of \ac{AI} systems. We developed the \ac{PHASE} guideline that translates these concepts for the analysis of \ac{AI} systems. 
Pilot testing the guidelines with students in a graduate-level responsible \ac{AI} course helped us validate and improve the final \ac{PHASE} guideline provided in the supplementary material. 
We find that \ac{PHASE} affords an analyst the ability to adopt a sociotechnical lens, establish traceability, and account for changes in a system over time. We invite the \ac{AI} community to leverage these affordances to address growing safety concerns with rapidly changing \ac{AI} systems, establish traceable \ac{AI} governance practices, and improve safety culture in the technology industry for the safety of all. 
\section{Generative AI Usage Statement}
The first author used ChatGPT (version 5.2) sparingly to obtain suggestions related to grammar, writing style, and conciseness. All suggestions were reviewed critically, and none were incorporated verbatim. Grammarly was additionally used for grammar and spelling edits. The remaining authors primarily provided feedback through verbal discussions and written comments on the manuscript and did not use generative AI tools for providing feedback.
\section{Ethical Considerations Statement}
In this study, we collected information at multiple stages of our methodology as described in Section 3 of the paper. In the information gathering stage for Case Study 1 and Case Study 2, we primarily relied on publicly available resources and consulted subject-matter experts as needed to clarify our understanding of the projects. As such, and following our institution’s research ethics board policy, research ethics approval for the two cases was not needed. Case Study 3, on the other hand, involved a qualitative study where we recruited and conducted semi-structured interviews with relevant stakeholders along the development process.  As this pertains to designed interventions with collection from human participants, we followed and received the necessary institutional approvals prior to conducting the study. All the participants (15 for the workshop and 8 for interviews) for Case Study 3 provided informed consent. 
\bibliographystyle{ACM-Reference-Format}
\bibliography{main}

@String{Computing = "Computing" }

@String{Computer = "{IEEE} Computer" }

@String{Springer = "Springer-Verlag" }

@ARTICLE{Thomasen2023-cc,
  title    = "{SAFETY} {IN} {ARTIFICIAL} {INTELLIGENCE} \& {ROBOTICS}
              {GOVERNANCE} {IN} {CANADA}",
  author   = "Thomasen, Kristen",
  journal  = "Bar Review",
  volume   =  101,
  number   =  1,
  month    =  may,
  year     =  2023,
  language = "en"
}

@MISC{The_White_House2023-ad,
  title        = "Executive Order on the Safe, Secure, and Trustworthy
                  Development and Use of Artificial Intelligence",
  author       = "{The White House}",
  month        =  oct,
  year         =  2023,
  howpublished = "\url{https://www.whitehouse.gov/briefing-room/presidential-actions/2023/10/30/executive-order-on-the-safe-secure-and-trustworthy-development-and-use-of-artificial-intelligence/}",
  note         = "Accessed: 2023-12-27"
}

@INPROCEEDINGS{Shelby2023-to,
  title     = "Sociotechnical Harms of Algorithmic Systems: Scoping a Taxonomy
               for Harm Reduction",
  booktitle = "Proceedings of the 2023 {AAAI/ACM} Conference on {AI}, Ethics,
               and Society",
  author    = "Shelby, Renee and Rismani, Shalaleh and Henne, Kathryn and Moon,
               Ajung and Rostamzadeh, Negar and Nicholas, Paul and
               Yilla-Akbari, N'mah and Gallegos, Jess and Smart, Andrew and
               Garcia, Emilio and Virk, Gurleen",
   publisher = "Association for Computing Machinery",
  pages     = "723--741",
  series    = "AIES '23",
  month     =  aug,
  year      =  2023,
  address   = "New York, NY, USA",
  location  = "Montr\textbackslash'\{e\}al, QC, Canada"
}

@INPROCEEDINGS{Chan2023-rk,
  title     = "Harms from Increasingly Agentic Algorithmic Systems",
  booktitle = "Proceedings of the 2023 {ACM} Conference on Fairness,
               Accountability, and Transparency",
  author    = "Chan, Alan and Salganik, Rebecca and Markelius, Alva and Pang,
               Chris and Rajkumar, Nitarshan and Krasheninnikov, Dmitrii and
               Langosco, Lauro and He, Zhonghao and Duan, Yawen and Carroll,
               Micah and Lin, Michelle and Mayhew, Alex and Collins, Katherine
               and Molamohammadi, Maryam and Burden, John and Zhao, Wanru and
               Rismani, Shalaleh and Voudouris, Konstantinos and Bhatt, Umang
               and Weller, Adrian and Krueger, David and Maharaj, Tegan",
  publisher = "Association for Computing Machinery",
  pages     = "651--666",
  series    = "FAccT '23",
  month     =  jun,
  year      =  2023,
  address   = "New York, NY, USA",
  location  = "Chicago, IL, USA"
}

@INPROCEEDINGS{Weidinger2022-ni,
  title     = "Taxonomy of Risks posed by Language Models",
  booktitle = "2022 {ACM} Conference on Fairness, Accountability, and
               Transparency",
  author    = "Weidinger, Laura and Uesato, Jonathan and Rauh, Maribeth and
               Griffin, Conor and Huang, Po-Sen and Mellor, John and Glaese,
               Amelia and Cheng, Myra and Balle, Borja and Kasirzadeh, Atoosa
               and Biles, Courtney and Brown, Sasha and Kenton, Zac and
               Hawkins, Will and Stepleton, Tom and Birhane, Abeba and
               Hendricks, Lisa Anne and Rimell, Laura and Isaac, William and
               Haas, Julia and Legassick, Sean and Irving, Geoffrey and
               Gabriel, Iason",
 
  publisher = "Association for Computing Machinery",
  pages     = "214--229",
  series    = "FAccT '22",
  month     =  jun,
  year      =  2022,
  address   = "New York, NY, USA",
  location  = "Seoul, Republic of Korea"
}

@ARTICLE{Karizat2021-ch,
  title     = "Algorithmic Folk Theories and Identity: How {TikTok} Users
               {Co-Produce} Knowledge of Identity and Engage in Algorithmic
               Resistance",
  author    = "Karizat, Nadia and Delmonaco, Dan and Eslami, Motahhare and
               Andalibi, Nazanin",
 
  journal   = "Proc. ACM Hum.-Comput. Interact.",
  publisher = "Association for Computing Machinery",
  volume    =  5,
  number    = "CSCW2",
  pages     = "1--44",
  month     =  oct,
  year      =  2021,
  address   = "New York, NY, USA"
}

@ARTICLE{Rakova2021-dg,
  title     = "Where Responsible {AI} meets Reality: Practitioner Perspectives
               on Enablers for Shifting Organizational Practices",
  author    = "Rakova, Bogdana and Yang, Jingying and Cramer, Henriette and
               Chowdhury, Rumman",
  journal   = "Proc. ACM Hum.-Comput. Interact.",
  publisher = "Association for Computing Machinery",
  volume    =  5,
  number    = "CSCW1",
  pages     = "1--23",
  month     =  apr,
  year      =  2021,
  address   = "New York, NY, USA"
}

@ARTICLE{Wong2023-nx,
  title     = "Seeing Like a Toolkit: How Toolkits Envision the Work of {AI}
               Ethics",
  author    = "Wong, Richmond Y and Madaio, Michael A and Merrill, Nick",
  journal   = "Proc. ACM Hum.-Comput. Interact.",
  publisher = "Association for Computing Machinery",
  volume    =  7,
  number    = "CSCW1",
  pages     = "1--27",
  month     =  apr,
  year      =  2023,
  address   = "New York, NY, USA"
}

@INPROCEEDINGS{Rismani2023-xt,
  title     = "From Plane Crashes to Algorithmic Harm: Applicability of Safety
               Engineering Frameworks for Responsible {ML}",
  booktitle = "Proceedings of the 2023 {CHI} Conference on Human Factors in
               Computing Systems",
  author    = "Rismani, Shalaleh and Shelby, Renee and Smart, Andrew and Jatho,
               Edgar and Kroll, Joshua and Moon, Ajung and Rostamzadeh, Negar",
 
  publisher = "Association for Computing Machinery",
  number    = "Article 2",
  pages     = "1--18",
  series    = "CHI '23",
  month     =  apr,
  year      =  2023,
  address   = "New York, NY, USA",
  location  = "Hamburg, Germany"
}

@ARTICLE{Madaio2022-vt,
  title     = "Assessing the Fairness of {AI} Systems: {AI} Practitioners'
               Processes, Challenges, and Needs for Support",
  author    = "Madaio, Michael and Egede, Lisa and Subramonyam, Hariharan and
               Wortman Vaughan, Jennifer and Wallach, Hanna",
  journal   = "Proc. ACM Hum.-Comput. Interact.",
  publisher = "Association for Computing Machinery",
  volume    =  6,
  number    = "CSCW1",
  pages     = "1--26",
  month     =  apr,
  year      =  2022,
  address   = "New York, NY, USA"
}

@ARTICLE{Jatho2023-zr,
  title         = "Concrete Safety for {ML} Problems: System Safety for {ML}
                   Development and Assessment",
  author        = "Jatho, Edgar W and Mailloux, Logan O and Williams, Eugene D
                   and McClure, Patrick and Kroll, Joshua A",
  month         =  feb,
  year          =  2023,
  archivePrefix = "arXiv",
  primaryClass  = "cs.LG",
  eprint        = "2302.02972"
}

@BOOK{Leveson2011-fo,
  title     = "Engineering a Safer World: Systems Thinking Applied to Safety",
  author    = "Leveson, Nancy",
  publisher = "MIT Press",
  year      =  2011,
  language  = "en"
}

@INPROCEEDINGS{Dobbe2022-ql,
  title     = "System Safety and Artificial Intelligence",
  booktitle = "2022 {ACM} Conference on Fairness, Accountability, and
               Transparency",
  author    = "Dobbe, Roel",
  publisher = "Association for Computing Machinery",
  pages     = "1584",
  series    = "FAccT '22",
  month     =  jun,
  year      =  2022,
  address   = "New York, NY, USA",
  location  = "Seoul, Republic of Korea"
}

@ARTICLE{Leveson1986-ag,
  title     = "Software safety: why, what, and how",
  author    = "Leveson, Nancy G",
  journal   = "ACM Comput. Surv.",
  publisher = "Association for Computing Machinery",
  volume    =  18,
  number    =  2,
  pages     = "125--163",
  month     =  jun,
  year      =  1986,
  address   = "New York, NY, USA"
}

@ARTICLE{Lazar2023-du,
  title    = "{AI} safety on whose terms?",
  author   = "Lazar, Seth and Nelson, Alondra",
  journal  = "Science",
  volume   =  381,
  number   =  6654,
  pages    = "138",
  month    =  jul,
  year     =  2023,
  language = "en"
}

@TECHREPORT{Leveson_undated-zz,
  title  = "{STPA} Handbook",
  author = "Leveson, Nancy and Thomas, John",
  month  =  mar,
  year   =  2018
}

@ARTICLE{Mantymaki2022-im,
  title         = "Putting {AI} Ethics into Practice: The Hourglass Model of
                   Organizational {AI} Governance",
  author        = "M{\"a}ntym{\"a}ki, Matti and Minkkinen, Matti and Birkstedt,
                   Teemu and Viljanen, Mika",
  month         =  jun,
  year          =  2022,
  archivePrefix = "arXiv",
  primaryClass  = "cs.AI",
  eprint        = "2206.00335"
}

@ARTICLE{Widder2022-rf,
  title         = "Dislocated Accountabilities in the {AI} Supply Chain:
                   Modularity and Developers' Notions of Responsibility",
  author        = "Widder, David Gray and Nafus, Dawn",
  month         =  sep,
  year          =  2022,
  archivePrefix = "arXiv",
  primaryClass  = "cs.CY",
  eprint        = "2209.09780"
}

@ARTICLE{Figueras2022-dm,
  title     = "Exploring tensions in Responsible {AI} in practice. An interview
               study on {AI} practices in and for Swedish public organizations",
  author    = "Figueras, Cl{\`a}udia and Verhagen, Harko and Pargman, Teresa
               Cerratto",
  journal   = "Scandinavian Journal of Information Systems",
  publisher = "aisel.aisnet.org",
  volume    =  34,
  number    =  2,
  pages     = "6",
  year      =  2022
}

@MISC{meta-system-card,
  title        = "System Cards, a new resource for understanding how {AI}
                  systems work",
  author = "Meta",  
  howpublished = "\url{https://ai.meta.com/blog/system-cards-a-new-resource-for-understanding-how-ai-systems-work/}",
  note         = "Accessed: 2024-8-8",
  language     = "en"
}

@ARTICLE{Nabavi2023-ce,
  title     = "Leverage zones in Responsible {AI}: towards a systems thinking
               conceptualization",
  author    = "Nabavi, Ehsan and Browne, Chris",
  journal   = "Humanities and Social Sciences Communications",
  publisher = "Palgrave",
  volume    =  10,
  number    =  1,
  pages     = "1--9",
  month     =  mar,
  year      =  2023,
  language  = "en"
}

@INPROCEEDINGS{Sloane2022-ag,
  title     = "German {AI} {Start-Ups} and {``AI} Ethics'': Using A Social
               Practice Lens for Assessing and Implementing {Socio-Technical}
               Innovation",
  booktitle = "2022 {ACM} Conference on Fairness, Accountability, and
               Transparency",
  author    = "Sloane, Mona and Zakrzewski, Janina",
  publisher = "Association for Computing Machinery",
  pages     = "935--947",
  series    = "FAccT '22",
  month     =  jun,
  year      =  2022,
  address   = "New York, NY, USA",
  location  = "Seoul, Republic of Korea"
}

@TECHREPORT{Khlaaf2023-qs,
  title       = "Toward Comprehensive Risk Assessments and Assurance of
                 {AI-Based} Systems",
  author      = "Khlaaf, Heidy",
  institution = "Trail of Bits",
  month       =  mar,
  year        =  2023
}

@ARTICLE{Brundage2020-ck,
  title         = "Toward Trustworthy {AI} Development: Mechanisms for
                   Supporting Verifiable Claims",
  author        = "Brundage, Miles and Avin, Shahar and Wang, Jasmine and
                   Belfield, Haydn and Krueger, Gretchen and Hadfield, Gillian
                   and Khlaaf, Heidy and Yang, Jingying and Toner, Helen and
                   Fong, Ruth and Maharaj, Tegan and Koh, Pang Wei and Hooker,
                   Sara and Leung, Jade and Trask, Andrew and Bluemke, Emma and
                   Lebensold, Jonathan and O'Keefe, Cullen and Koren, Mark and
                   Ryffel, Th{\'e}o and Rubinovitz, J B and Besiroglu, Tamay
                   and Carugati, Federica and Clark, Jack and Eckersley, Peter
                   and de Haas, Sarah and Johnson, Maritza and Laurie, Ben and
                   Ingerman, Alex and Krawczuk, Igor and Askell, Amanda and
                   Cammarota, Rosario and Lohn, Andrew and Krueger, David and
                   Stix, Charlotte and Henderson, Peter and Graham, Logan and
                   Prunkl, Carina and Martin, Bianca and Seger, Elizabeth and
                   Zilberman, Noa and h{\'E}igeartaigh, Se{\'a}n {\'O} and
                   Kroeger, Frens and Sastry, Girish and Kagan, Rebecca and
                   Weller, Adrian and Tse, Brian and Barnes, Elizabeth and
                   Dafoe, Allan and Scharre, Paul and Herbert-Voss, Ariel and
                   Rasser, Martijn and Sodhani, Shagun and Flynn, Carrick and
                   Gilbert, Thomas Krendl and Dyer, Lisa and Khan, Saif and
                   Bengio, Yoshua and Anderljung, Markus",
  month         =  apr,
  year          =  2020,
  archivePrefix = "arXiv",
  primaryClass  = "cs.CY",
  eprint        = "2004.07213"
}

@ARTICLE{Yu2022-ed,
  title         = "Scaling Autoregressive Models for {Content-Rich}
                   {Text-to-Image} Generation",
  author        = "Yu, Jiahui and Xu, Yuanzhong and Koh, Jing Yu and Luong,
                   Thang and Baid, Gunjan and Wang, Zirui and Vasudevan, Vijay
                   and Ku, Alexander and Yang, Yinfei and Ayan, Burcu Karagol
                   and Hutchinson, Ben and Han, Wei and Parekh, Zarana and Li,
                   Xin and Zhang, Han and Baldridge, Jason and Wu, Yonghui",
  month         =  jun,
  year          =  2022,
  archivePrefix = "arXiv",
  primaryClass  = "cs.CV",
  eprint        = "2206.10789"
}

@ARTICLE{Franceschelli2021-kj,
  title         = "Creativity and Machine Learning: A Survey",
  author        = "Franceschelli, Giorgio and Musolesi, Mirco",
  month         =  apr,
  year          =  2021,
  archivePrefix = "arXiv",
  primaryClass  = "cs.LG",
  eprint        = "2104.02726"
}

@ARTICLE{Saharia2022-em,
  title         = "Photorealistic {Text-to-Image} Diffusion Models with Deep
                   Language Understanding",
  author        = "Saharia, Chitwan and Chan, William and Saxena, Saurabh and
                   Li, Lala and Whang, Jay and Denton, Emily and Ghasemipour,
                   Seyed Kamyar Seyed and Ayan, Burcu Karagol and Sara Mahdavi,
                   S and Lopes, Rapha Gontijo and Salimans, Tim and Ho,
                   Jonathan and Fleet, David J and Norouzi, Mohammad",
  month         =  may,
  year          =  2022,
  archivePrefix = "arXiv",
  primaryClass  = "cs.CV",
  eprint        = "2205.11487"
}

@ARTICLE{Van_den_Berg2023-oy,
  title    = "Development and clinical impact assessment of a machine-learning
              model for early prediction of late-onset sepsis",
  author   = "van den Berg, Merel A M and Medina, O'jay O A G and Loohuis,
              Ingmar I P and van der Flier, Michiel M and Dudink, Jeroen J and
              Benders, Manon M J N L and Bartels, Richard R T and Vijlbrief,
              Daniel D C",
  journal  = "Comput. Biol. Med.",
  volume   =  163,
  pages    = "107156",
  month    =  jun,
  year     =  2023,
  language = "en"
}

@ARTICLE{Njie2014-ae,
  title     = "Case study as a choice in qualitative methodology",
  author    = "Njie, Baboucarr and Asimiran, Soaib",
  journal   = "IOSR J. Res. Method Educ. (IOSRJRME)",
  publisher = "IOSR Journals",
  volume    =  4,
  number    =  3,
  pages     = "35--40",
  year      =  2014,
  language  = "en"
}

@ARTICLE{Ganguli2022-qn,
  title         = "Red Teaming Language Models to Reduce Harms: Methods,
                   Scaling Behaviors, and Lessons Learned",
  author        = "Ganguli, Deep and Lovitt, Liane and Kernion, Jackson and
                   Askell, Amanda and Bai, Yuntao and Kadavath, Saurav and
                   Mann, Ben and Perez, Ethan and Schiefer, Nicholas and
                   Ndousse, Kamal and Jones, Andy and Bowman, Sam and Chen,
                   Anna and Conerly, Tom and DasSarma, Nova and Drain, Dawn and
                   Elhage, Nelson and El-Showk, Sheer and Fort, Stanislav and
                   Hatfield-Dodds, Zac and Henighan, Tom and Hernandez, Danny
                   and Hume, Tristan and Jacobson, Josh and Johnston, Scott and
                   Kravec, Shauna and Olsson, Catherine and Ringer, Sam and
                   Tran-Johnson, Eli and Amodei, Dario and Brown, Tom and
                   Joseph, Nicholas and McCandlish, Sam and Olah, Chris and
                   Kaplan, Jared and Clark, Jack",
  month         =  aug,
  year          =  2022,
  archivePrefix = "arXiv",
  primaryClass  = "cs.CL",
  eprint        = "2209.07858"
}

@INPROCEEDINGS{Raji2020-aq,
  title     = "Closing the {AI} accountability gap: defining an end-to-end
               framework for internal algorithmic auditing",
  booktitle = "Proceedings of the 2020 Conference on Fairness, Accountability,
               and Transparency",
  author    = "Raji, Inioluwa Deborah and Smart, Andrew and White, Rebecca N
               and Mitchell, Margaret and Gebru, Timnit and Hutchinson, Ben and
               Smith-Loud, Jamila and Theron, Daniel and Barnes, Parker",
  publisher = "Association for Computing Machinery",
  pages     = "33--44",
  series    = "FAT* '20",
  month     =  jan,
  year      =  2020,
  address   = "New York, NY, USA",
  location  = "Barcelona, Spain"
}

@ARTICLE{Li2022-wj,
  title    = "{FMEA-AI}: {AI} fairness impact assessment using failure mode and
              effects analysis",
  author   = "Li, Jamy and Chignell, Mark",
  journal  = "AI and Ethics",
  volume   =  2,
  number   =  4,
  pages    = "837--850",
  month    =  nov,
  year     =  2022
}

@ARTICLE{Rismani2021-gr,
  title     = "How do {AI} systems fail socially?: an engineering risk analysis
               approach",
  author    = "Rismani, S and Moon, A",
  journal   = "on Ethics in Engineering, Science and …",
  publisher = "ieeexplore.ieee.org",
  year      =  2021
}

@INPROCEEDINGS{Watkins2021-wc,
  title     = "Governing Algorithmic Systems with Impact Assessments: Six
               Observations",
  booktitle = "Proceedings of the 2021 {AAAI/ACM} Conference on {AI}, Ethics,
               and Society",
  author    = "Watkins, Elizabeth Anne and Moss, Emanuel and Metcalf, Jacob and
               Singh, Ranjit and Elish, Madeleine Clare",
  
  publisher = "Association for Computing Machinery",
  pages     = "1010--1022",
  series    = "AIES '21",
  month     =  jul,
  year      =  2021,
  address   = "New York, NY, USA",
  location  = "Virtual Event, USA"
}

@ARTICLE{Mantelero2021-cr,
  title     = "An evidence-based methodology for human rights impact assessment
               ({HRIA}) in the development of {AI} data-intensive systems",
  author    = "Mantelero, Alessandro and Esposito, Maria Samantha",
  
  journal   = "Computer Law \& Security Review",
  publisher = "Elsevier",
  volume    =  41,
  pages     = "105561",
  month     =  jul,
  year      =  2021
}

@ARTICLE{Reisman2018-fs,
  title     = "Algorithmic Impact Assessments: A Practical Framework for Public
               Agency",
  author    = "Reisman, Dillon and Schultz, Jason and Crawford, Kate and
               Whittaker, Meredith",
  journal   = "AI Now",
  publisher = "nist.gov",
  year      =  2018
}

@MISC{ODSC-Open_Data_Science2022-re,
  title        = "15 Open Source Responsible {AI} Toolkits and Projects to Use
                  Today",
  booktitle    = "Medium",
  author       = "{ODSC-Open Data Science}",
  month        =  jan,
  year         =  2022,
  howpublished = "\url{https://odsc.medium.com/15-open-source-responsible-ai-toolkits-and-projects-to-use-today-fbc1c2ea2815}",
  note         = "Accessed: 2023-12-27",
  language     = "en"
}

@INPROCEEDINGS{Barocas2021-sr,
  title     = "Designing Disaggregated Evaluations of {AI} Systems: Choices,
               Considerations, and Tradeoffs",
  booktitle = "Proceedings of the 2021 {AAAI/ACM} Conference on {AI}, Ethics,
               and Society",
  author    = "Barocas, Solon and Guo, Anhong and Kamar, Ece and Krones,
               Jacquelyn and Morris, Meredith Ringel and Vaughan, Jennifer
               Wortman and Wadsworth, W Duncan and Wallach, Hanna",
 
  publisher = "Association for Computing Machinery",
  pages     = "368--378",
  series    = "AIES '21",
  month     =  jul,
  year      =  2021,
  address   = "New York, NY, USA",
  location  = "Virtual Event, USA"
}

@INPROCEEDINGS{Knight2002-dy,
  title     = "Safety critical systems: challenges and directions",
  booktitle = "Proceedings of the 24th International Conference on Software
               Engineering. {ICSE} 2002",
  author    = "Knight, J C",
  publisher = "IEEE",
  pages     = "547--550",
  year      =  2002
}

@BOOK{Swuste2022-xd,
  title     = "From Safety to Safety Science: The Evolution of Thinking and
               Practice",
  author    = "Swuste, Paul",
  publisher = "Routledge",
  year      =  2022,
  language  = "en"
}

@BOOK{Calder1899-gr,
  title  = "The Prevention Of Factory Accidents: Being An Account Of
            Manufacturing Industry And Accident",
  author = "Calder, John",
  year   =  1899
}

@INPROCEEDINGS{Bender2021-wy,
  title     = "On the Dangers of Stochastic Parrots: Can Language Models Be Too
               Big?",
  booktitle = "Proceedings of the 2021 {ACM} Conference on Fairness,
               Accountability, and Transparency",
  author    = "Bender, Emily M and Gebru, Timnit and McMillan-Major, Angelina
               and Shmitchell, Shmargaret",
  publisher = "Association for Computing Machinery",
  pages     = "610--623",
  series    = "FAccT '21",
  month     =  mar,
  year      =  2021,
  address   = "New York, NY, USA",
  location  = "Virtual Event, Canada"
}

@ARTICLE{Birhane2021-ry,
  title         = "Multimodal datasets: misogyny, pornography, and malignant
                   stereotypes",
  author        = "Birhane, Abeba and Prabhu, Vinay Uday and Kahembwe, Emmanuel",
  month         =  oct,
  year          =  2021,
  archivePrefix = "arXiv",
  primaryClass  = "cs.CY",
  eprint        = "2110.01963"
}

@INPROCEEDINGS{Solaiman2023-dp,
  title     = "The Gradient of Generative {AI} Release: Methods and
               Considerations",
  booktitle = "Proceedings of the 2023 {ACM} Conference on Fairness,
               Accountability, and Transparency",
  author    = "Solaiman, Irene",
  publisher = "Association for Computing Machinery",
  pages     = "111--122",
  series    = "FAccT '23",
  month     =  jun,
  year      =  2023,
  address   = "New York, NY, USA",
  location  = "Chicago, IL, USA"
}

@INPROCEEDINGS{Bucknall2022-hw,
  title     = "Current and {Near-Term} {AI} as a Potential Existential Risk
               Factor",
  booktitle = "Proceedings of the 2022 {AAAI/ACM} Conference on {AI}, Ethics,
               and Society",
  author    = "Bucknall, Benjamin S and Dori-Hacohen, Shiri",
  publisher = "Association for Computing Machinery",
  pages     = "119--129",
  series    = "AIES '22",
  month     =  jul,
  year      =  2022,
  address   = "New York, NY, USA",
  location  = "Oxford, United Kingdom"
}

@INCOLLECTION{Russell2022-hc,
  title     = "Artificial Intelligence and the Problem of Control",
  booktitle = "Perspectives on Digital Humanism",
  author    = "Russell, Stuart",
  editor    = "Werthner, Hannes and Prem, Erich and Lee, Edward A and Ghezzi,
               Carlo",
  publisher = "Springer International Publishing",
  pages     = "19--24",
  year      =  2022,
  address   = "Cham"
}

@ARTICLE{Vera_Liao2023-gv,
  title         = "Rethinking Model Evaluation as Narrowing the
                   {Socio-Technical} Gap",
  author        = "Vera Liao, Q and Xiao, Ziang",
  month         =  jun,
  year          =  2023,
  archivePrefix = "arXiv",
  primaryClass  = "cs.HC",
  eprint        = "2306.03100"
}

@INPROCEEDINGS{Birhane2022-vp,
  title       = "The values encoded in machine learning research",
  booktitle   = "2022 {ACM} Conference on Fairness, Accountability, and
                 Transparency",
  author      = "Birhane, Abeba and Kalluri, Pratyusha and Card, Dallas and
                 Agnew, William and Dotan, Ravit and Bao, Michelle",
  publisher   = "ACM",
  month       =  jun,
  year        =  2022,
  address     = "New York, NY, USA",
  conference  = "FAccT '22: 2022 ACM Conference on Fairness, Accountability,
                 and Transparency",
  location    = "Seoul Republic of Korea",
  original_id = "f4ef4c20-edae-0cf1-ae3c-4f25abcf85fa"
}

@INPROCEEDINGS{Yang2020-ha,
  title     = "Re-examining Whether, Why, and How {Human-AI} Interaction Is
               Uniquely Difficult to Design",
  booktitle = "Proceedings of the 2020 {CHI} Conference on Human Factors in
               Computing Systems",
  author    = "Yang, Qian and Steinfeld, Aaron and Ros{\'e}, Carolyn and
               Zimmerman, John",
  publisher = "Association for Computing Machinery",
  pages     = "1--13",
  series    = "CHI '20",
  month     =  apr,
  year      =  2020,
  address   = "New York, NY, USA",
  location  = "Honolulu, HI, USA"
}

@UNPUBLISHED{Ahmed2023-kv,
  title    = "Building the Epistemic Community of {AI} Safety",
  author   = "Ahmed, Shazeda and Ja{\'z}wi{\'n}ska, Klaudia and Ahlawat,
              Archana and Winecoff, Amy and Wang, Mona",
  month    =  nov,
  year     =  2023
}

@ARTICLE{Bai2022-ai,
  title         = "Constitutional {AI}: Harmlessness from {AI} Feedback",
  author        = "Bai, Yuntao and Kadavath, Saurav and Kundu, Sandipan and
                   Askell, Amanda and Kernion, Jackson and Jones, Andy and
                   Chen, Anna and Goldie, Anna and Mirhoseini, Azalia and
                   McKinnon, Cameron and Chen, Carol and Olsson, Catherine and
                   Olah, Christopher and Hernandez, Danny and Drain, Dawn and
                   Ganguli, Deep and Li, Dustin and Tran-Johnson, Eli and
                   Perez, Ethan and Kerr, Jamie and Mueller, Jared and Ladish,
                   Jeffrey and Landau, Joshua and Ndousse, Kamal and Lukosuite,
                   Kamile and Lovitt, Liane and Sellitto, Michael and Elhage,
                   Nelson and Schiefer, Nicholas and Mercado, Noemi and
                   DasSarma, Nova and Lasenby, Robert and Larson, Robin and
                   Ringer, Sam and Johnston, Scott and Kravec, Shauna and El
                   Showk, Sheer and Fort, Stanislav and Lanham, Tamera and
                   Telleen-Lawton, Timothy and Conerly, Tom and Henighan, Tom
                   and Hume, Tristan and Bowman, Samuel R and Hatfield-Dodds,
                   Zac and Mann, Ben and Amodei, Dario and Joseph, Nicholas and
                   McCandlish, Sam and Brown, Tom and Kaplan, Jared",
  month         =  dec,
  year          =  2022,
  archivePrefix = "arXiv",
  primaryClass  = "cs.CL",
  eprint        = "2212.08073"
}

@ARTICLE{Wenzel2023-pi,
  title         = "Can Voice Assistants Be Microaggressors? {Cross-Race}
                   Psychological Responses to Failures of Automatic Speech
                   Recognition",
  author        = "Wenzel, Kimi and Devireddy, Nitya and Davidson, Cam and
                   Kaufman, Geoff",
  month         =  feb,
  year          =  2023,
  archivePrefix = "arXiv",
  primaryClass  = "cs.HC",
  eprint        = "2302.12326"
}

@INPROCEEDINGS{Blodgett2022-db,
  title     = "Responsible Language Technologies: Foreseeing and Mitigating
               Harms",
  booktitle = "Extended Abstracts of the 2022 {CHI} Conference on Human Factors
               in Computing Systems",
  author    = "Blodgett, Su Lin and Liao, Q Vera and Olteanu, Alexandra and
               Mihalcea, Rada and Muller, Michael and Scheuerman, Morgan Klaus
               and Tan, Chenhao and Yang, Qian",
  publisher = "Association for Computing Machinery",
  number    = "Article 152",
  pages     = "1--3",
  series    = "CHI EA '22",
  month     =  apr,
  year      =  2022,
  address   = "New York, NY, USA",
  location  = "New Orleans, LA, USA"
}

@ARTICLE{Bhatt_Surabhi_undated-ye,
  title     = "Interpretable Machine Learning Models for Clinical
               {Decision-Making} in a {High-Need}, {Value-Based} Primary Care
               Setting",
  author    = "{Bhatt Surabhi} and {Cohon Adam} and {Rose Jenna} and {Majerczyk
               Natalia} and {Cozzi Brian} and {Crenshaw Drew} and {Myers
               Griffin}",
  journal   = "NEJM Catalyst",
  publisher = "Massachusetts Medical Society",
  volume    =  2,
  number    =  4, 
year = 2021
}

@ARTICLE{Lima2023-bs,
  title         = "Blaming Humans and Machines: What Shapes People's Reactions
                   to Algorithmic Harm",
  author        = "Lima, Gabriel and Grgi{\'c}-Hla{\v c}a, Nina and Cha,
                   Meeyoung",
  month         =  apr,
  year          =  2023,
  archivePrefix = "arXiv",
  primaryClass  = "cs.CY",
  eprint        = "2304.02176"
}

@ARTICLE{Leveson2004-jn,
  title    = "A new accident model for engineering safer systems",
  author   = "Leveson, Nancy",
  journal  = "Saf. Sci.",
  volume   =  42,
  number   =  4,
  pages    = "237--270",
  month    =  apr,
  year     =  2004
}

@ARTICLE{Von_Bertalanffy1972-uw,
  title     = "The History and Status of General Systems Theory",
  author    = "Von Bertalanffy, Ludwig",
  journal   = "Acad. Manage. J.",
  publisher = "Academy of Management",
  volume    =  15,
  number    =  4,
  pages     = "407--426",
  year      =  1972
}

@ARTICLE{Leveson2017-di,
  title    = "Rasmussen's legacy: A paradigm change in engineering for safety",
  author   = "Leveson, Nancy G",
  journal  = "Appl. Ergon.",
  volume   =  59,
  number   = "Pt B",
  pages    = "581--591",
  month    =  mar,
  year     =  2017,
  language = "en"
}

@MISC{Oecd2024-kr,
  title        = "{OECD} {AI} Principles overview",
  author       = "{OECD}",
  year         =  2024,
  howpublished = "\url{https://oecd.ai/en/ai-principles}",
  note         = "Accessed: 2024-1-21",
  language     = "en"
}

@BOOK{Carlson2012-iq,
  title     = "Effective {FMEAs}: Achieving Safe, Reliable, and Economical
               Products and Processes Using Failure Mode and Effects Analysis",
  author    = "Carlson, Carl",
  publisher = "Wiley \& Sons, Limited, John",
  year      =  2012
}

@BOOK{Crawford2022-qx,
  title     = "Atlas of {AI} Power, Politics, and the Planetary Costs of
               Artificial Intelligence",
  author    = "Crawford, Kate",
  publisher = "Yale University Press",
  month     =  aug,
  year      =  2022
}

@INPROCEEDINGS{Kroll2021-nk,
  title     = "Outlining Traceability: A Principle for Operationalizing
               Accountability in Computing Systems",
  booktitle = "Proceedings of the 2021 {ACM} Conference on Fairness,
               Accountability, and Transparency",
  author    = "Kroll, Joshua A",
  publisher = "Association for Computing Machinery",
  pages     = "758--771",
  series    = "FAccT '21",
  month     =  mar,
  year      =  2021,
  address   = "New York, NY, USA",
  location  = "Virtual Event, Canada"
}

@INPROCEEDINGS{DeVos2022-jk,
  title     = "Toward {User-Driven} Algorithm Auditing: Investigating users'
               strategies for uncovering harmful algorithmic behavior",
  booktitle = "Proceedings of the 2022 {CHI} Conference on Human Factors in
               Computing Systems",
  author    = "DeVos, Alicia and Dhabalia, Aditi and Shen, Hong and Holstein,
               Kenneth and Eslami, Motahhare",
  publisher = "dl.acm.org",
  pages     = "1--19",
  year      =  2022
}

@ARTICLE{Wang2022-cx,
  title     = "Understanding the Design Space of {AI-Mediated} Social
               Interaction in Online Learning: Challenges and Opportunities",
  author    = "Wang, Qiaosi and Camacho, Ida and Jing, Shan and Goel, Ashok K",
  journal   = "Proc. ACM Hum.-Comput. Interact.",
  publisher = "Association for Computing Machinery",
  volume    =  6,
  number    = "CSCW1",
  pages     = "1--26",
  month     =  apr,
  year      =  2022,
  address   = "New York, NY, USA"
}

@MISC{Shavit_undated-tf,
  title        = "Practices for governing agentic {AI} systems",
  author       = "Shavit, Yonadav and Agarwal, Sandhini and Brundage, Miles and
                  Adler Cullen O'keefe, Steven and Campbell, Rosie and Lee,
                  Teddy and Mishkin, Pamela and Eloundou, Tyna and Hickey, Alan
                  and Slama, Katarina and Ahmad, Lama and Mcmillan, Paul and
                  Beutel, Alex and Robinson, David G",
  howpublished = "\url{https://cdn.openai.com/papers/practices-for-governing-agentic-ai-systems.pdf}",
  note         = "Accessed: 2024-1-4", 
year = 2023
}

@MISC{Selbst2021-pk,
  title        = "An Institutional View of Algorithmic Assessments",
  author       = "Selbst, Andrew D",
  year         =  2021,
  howpublished = "\url{https://jolt.law.harvard.edu/assets/articlePDFs/v35/Selbst-An-Institutional-View-of-Algorithmic-Impact-Assessments.pdf}",
  note         = "Accessed: 2023-2-10"
}

@ARTICLE{Smuha2021-cp,
  title     = "Beyond the individual: governing {AI's} societal harm",
  author    = "Smuha, Nathalie A",
  journal   = "Internet Pol. Rev.",
  publisher = "Internet Policy Review, Alexander von Humboldt Institute for
               Internet and Society",
  volume    =  10,
  number    =  3,
  month     =  sep,
  year      =  2021,
  language  = "en"
}

@TECHREPORT{Moss2020-xl,
  title       = "Ethics Owners A New Model of Organizational Responsibility in
                 {Data-Driven} Technology Companies",
  author      = "Moss, Emanuel and Metcalf, Jacob",
  institution = "Data and Society",
  month       =  sep,
  year        =  2020
}

@ARTICLE{Weidinger2023-pe,
  title         = "Sociotechnical Safety Evaluation of Generative {AI} Systems",
  author        = "Weidinger, Laura and Rauh, Maribeth and Marchal, Nahema and
                   Manzini, Arianna and Hendricks, Lisa Anne and Mateos-Garcia,
                   Juan and Bergman, Stevie and Kay, Jackie and Griffin, Conor
                   and Bariach, Ben and Gabriel, Iason and Rieser, Verena and
                   Isaac, William",
  month         =  oct,
  year          =  2023,
  archivePrefix = "arXiv",
  primaryClass  = "cs.AI",
  eprint        = "2310.11986"
}

@MISC{openAIsafetypolicy,
  title        = "Safety standards",
  howpublished = "\url{https://openai.com/safety-standards}",
author = OpenAI, 
  note         = "Accessed: 2024-1-23",
  language     = "en", 
year = 2023
}

@ARTICLE{Hanna_undated-wm,
  title    = "{AI} Causes Real Harm. Let's Focus on That over the
              {End-of-Humanity} Hype",
  author   = "Hanna, Alex and Bender, Emily M",
  journal  = "Scientific American",
  language = "en",
year = 2023
}

@ARTICLE{Saetra2023-jq,
  title    = "Resolving the battle of short- vs. long-term {AI} risks",
  author   = "S{\ae}tra, Henrik Skaug and Danaher, John",
  journal  = "AI and Ethics",
  month    =  sep,
  year     =  2023
}

@INPROCEEDINGS{Wang2022-ck,
  title     = "Measuring Representational Harms in Image Captioning",
  booktitle = "2022 {ACM} Conference on Fairness, Accountability, and
               Transparency",
  author    = "Wang, Angelina and Barocas, Solon and Laird, Kristen and
               Wallach, Hanna",
  publisher = "Association for Computing Machinery",
  pages     = "324--335",
  series    = "FAccT '22",
  month     =  jun,
  year      =  2022,
  address   = "New York, NY, USA",
  location  = "Seoul, Republic of Korea"
}

@MISC{euAIact,
  title        = "Artificial Intelligence Act: deal on comprehensive rules for
                  trustworthy {AI}",
  month        =  sep,
  year         =  2023,
  howpublished = "\url{https://www.europarl.europa.eu/news/en/press-room/20231206IPR15699/artificial-intelligence-act-deal-on-comprehensive-rules-for-trustworthy-ai}",
  note         = "Accessed: 2024-1-22",
  language     = "en"
}

@ARTICLE{Basu2023-er,
  title         = "On the Challenges of using Reinforcement Learning in
                   Precision Drug Dosing: Delay and Prolongedness of Action
                   Effects",
  author        = "Basu, Sumana and Legault, Marc-Andr{\'e} and Romero-Soriano,
                   Adriana and Precup, Doina",
  month         =  jan,
  year          =  2023,
  archivePrefix = "arXiv",
  primaryClass  = "cs.LG",
  eprint        = "2301.00512"
}

@inproceedings{bird2023,
author = {Bird, Charlotte and Ungless, Eddie and Kasirzadeh, Atoosa},
title = {Typology of Risks of Generative Text-to-Image Models},
year = {2023},
isbn = {9798400702310},
publisher = {Association for Computing Machinery},
address = {New York, NY, USA},
url = {https://doi.org/10.1145/3600211.3604722},
doi = {10.1145/3600211.3604722},
booktitle = {Proceedings of the 2023 AAAI/ACM Conference on AI, Ethics, and Society},
pages = {396–410},
numpages = {15},
location = {<conf-loc>, <city>Montr\'{e}al</city>, <state>QC</state>, <country>Canada</country>, </conf-loc>},
series = {AIES '23}
}

@MISC{GPTsystemcard,
  title        = "{GPT}-5 System Card",
  author       = "Open AI",
  year  = "2025", 
  howpublished = "\url{https://openai.com/index/gpt-5-system-card/}",
  note         = "Accessed: 2025-8-10",
  language     = "en"
}

@INPROCEEDINGS{Cobbe2023-eg,
  title     = "Understanding accountability in algorithmic supply chains",
  author    = "Cobbe, Jennifer and Veale, Michael and Singh, Jatinder",
  booktitle = "2023 ACM Conference on Fairness Accountability and Transparency",
  publisher = "ACM",
  address   = "New York, NY, USA",
  pages     = "1186--1197",
  series    = "FAccT '23",
  month     =  jun,
  year      =  2023
}

@ARTICLE{Gyevnar2025-uw,
  title     = "{AI} safety for everyone",
  author    = "Gyevnár, Bálint and Kasirzadeh, Atoosa",
  journal   = "Nat. Mach. Intell.",
  publisher = "Springer Science and Business Media LLC",
  volume    =  7,
  number    =  4,
  pages     = "531--542",
  month     =  apr,
  year      =  2025,
  language  = "en"
}

@techreport{Bengio2025InternationalAISafetyReport,
  title        = {International AI Safety Report},
  author       = {Bengio, Yoshua and Mindermann, S{\"o}ren and Privitera, Daniele and Besiroglu, Tamay and Bommasani, Rishi and Casper, Stephen and Choi, Yejin and Fox, Peter and Garfinkel, Ben and Goldfarb, David and Heidari, Hoda and Ho, Andrew and Kapoor, Saurabh and Khalatbari, Leila and Longpre, Shayne and Manning, Christopher D. and Mavroudis, Vasilios and Mazeika, Mantas and Michael, Julian and Newman, Jessica and Ng, Kelvin Y. and Okolo, Chinasa T. and Raji, Inioluwa Deborah and Sastry, Girish and Seger, Elizabeth and Skeadas, Theodoros and South, Thomas and Strubell, Emma and Tram{\`e}r, Florian and Velasco, Luis and Wheeler, Nathan and Acemoglu, Daron and Adekanmbi, Olufunke and Dalrymple, David and Dietterich, Thomas G. and Fung, Pascale and Gourinchas, Pierre-Olivier and Heintz, Fredrik and Hinton, Geoffrey and Jennings, Nicholas and Krause, Andreas and Leavy, Susan and Liang, Percy and Ludermir, Teresa and Marda, Vidushi and Margetts, Helen and McDermid, John and Munga, John and Narayanan, Arvind and Nelson, Alondra and Neppel, Christopher and Oh, Alice and Ramchurn, Sarvapali and Russell, Stuart and Schaake, Marietje and Sch{\"o}lkopf, Bernhard and Song, Dawn and Soto, Andres and Tiedrich, Lukas and Varoquaux, Ga{\"e}l and Felten, Edward W. and Yao, Andrew and Zhang, Yong-Qiang and Ajala, Olugbenga and Albalawi, Fahad and Alserkal, Mayada and Avrin, Gabriel and Busch, Christoph and de Carvalho, Antonio C. P. L. F. and Fox, Benjamin and Gill, Amanpreet S. and Hatip, Ahmet H. and Heikkil{\"a}, Jussi and Johnson, Colin and Jolly, Gavin and Katzir, Zohar and Khan, Shakir M. and Kitano, Hiroaki and Kr{\"u}ger, Antonio and Lee, Kwan Min and Ligot, Daniel V. and L{\'o}pez Portillo, Jos{\'e} R. and Molchanovskyi, Oleksii and Monti, Andrea and Mwamanzi, Nothando and Nemer, Myriam and Oliver, Nuria and Pezoa Rivera, Rodrigo and Ravindran, Balaraman and Riza, Hasan and Rugege, Celestin and Seoighe, Cathal and Sheikh, Huma and Sheehan, Jack and Wong, David and Zeng, Yi},
  institution  = {UK Department for Science, Innovation and Technology},
  number       = {DSIT 2025/001},
  year         = {2025},
  url          = {https://www.gov.uk/government/publications/international-ai-safety-report-2025},
  note         = {Accessed: 2025-01-13}
}

@ARTICLE{Attard-Frost2025-tv,
  title     = "The ethics of {AI} value chains",
  author    = "Attard-Frost, Blair and Widder, David Gray",
  journal   = "Big Data Soc.",
  publisher = "SAGE Publications",
  volume    =  12,
  number    =  2,
  month     =  jun,
  year      =  2025,
  language  = "en"
}

@ARTICLE{Bas2020-dw,
  title     = "{STPA} methodology in a socio-technical system of monitoring and
               tracking diabetes mellitus",
  author    = "Bas, Esra",
  journal   = "Appl. Ergon.",
  publisher = "Elsevier",
  volume    =  89,
  pages     = "103190",
  month     =  nov,
  year      =  2020,
  language  = "en"
}

@INPROCEEDINGS{Chen2020-cy,
  title     = "Identifying Accident Causes of {Driver-Vehicle} Interactions
               Using System Theoretic Process Analysis ({STPA})",
  booktitle = "2020 {IEEE} International Conference on Systems, Man, and
               Cybernetics ({SMC})",
  author    = "Chen, Shufeng and Khastgir, Siddartha and Babaev, Islam and
               Jennings, Paul",
  publisher = "ieeexplore.ieee.org",
  pages     = "3247--3253",
  month     =  oct,
  year      =  2020
}

@misc{EUAIActArticle3,
  title        = {Article 3: Definitions, including definition of “AI system”},
  author       = {{European Union}},
  howpublished = {\url{https://artificialintelligenceact.eu/article/3/}},
  note         = {Accessed: 2025-01-13. According to Article 3(1), an “AI system” means “a machine-based system that is designed to operate with varying levels of autonomy and that may exhibit adaptiveness after deployment, and that, for explicit or implicit objectives, infers, from the input it receives, how to generate outputs such as predictions, content, recommendations, or decisions that can influence physical or virtual environments.”},
  year         = {2024}
}

@INPROCEEDINGS{Park2022-qr,
  title     = "A Study for {STPA-based} Identification of Safety Requirements
               from the Perspective of Drivers in {Take-Over} Request",
  booktitle = "2022 {IEEE} 3rd International Conference on {Human-Machine}
               Systems ({ICHMS})",
  author    = "Park, Jongwoo and Lee, Myeongkyu and Maeng, Jooyoung and Ahn,
               Changnam and Yang, Ji Hyun",
  publisher = "ieeexplore.ieee.org",
  pages     = "1--3",
  month     =  nov,
  year      =  2022
}

@TECHREPORT{Suo2017-hc,
  title       = "Integrating {STPA} into {ISO} 26262 process for requirement
                 development",
  author      = "Suo, Dajiang and Yako, Sarra and Boesch, Mathew and Post, Kyle",
  institution = "SAE Technical Paper",
  year        =  2017
}

\appendix

\section{Supplementary Material}

\includepdf[pages=-]{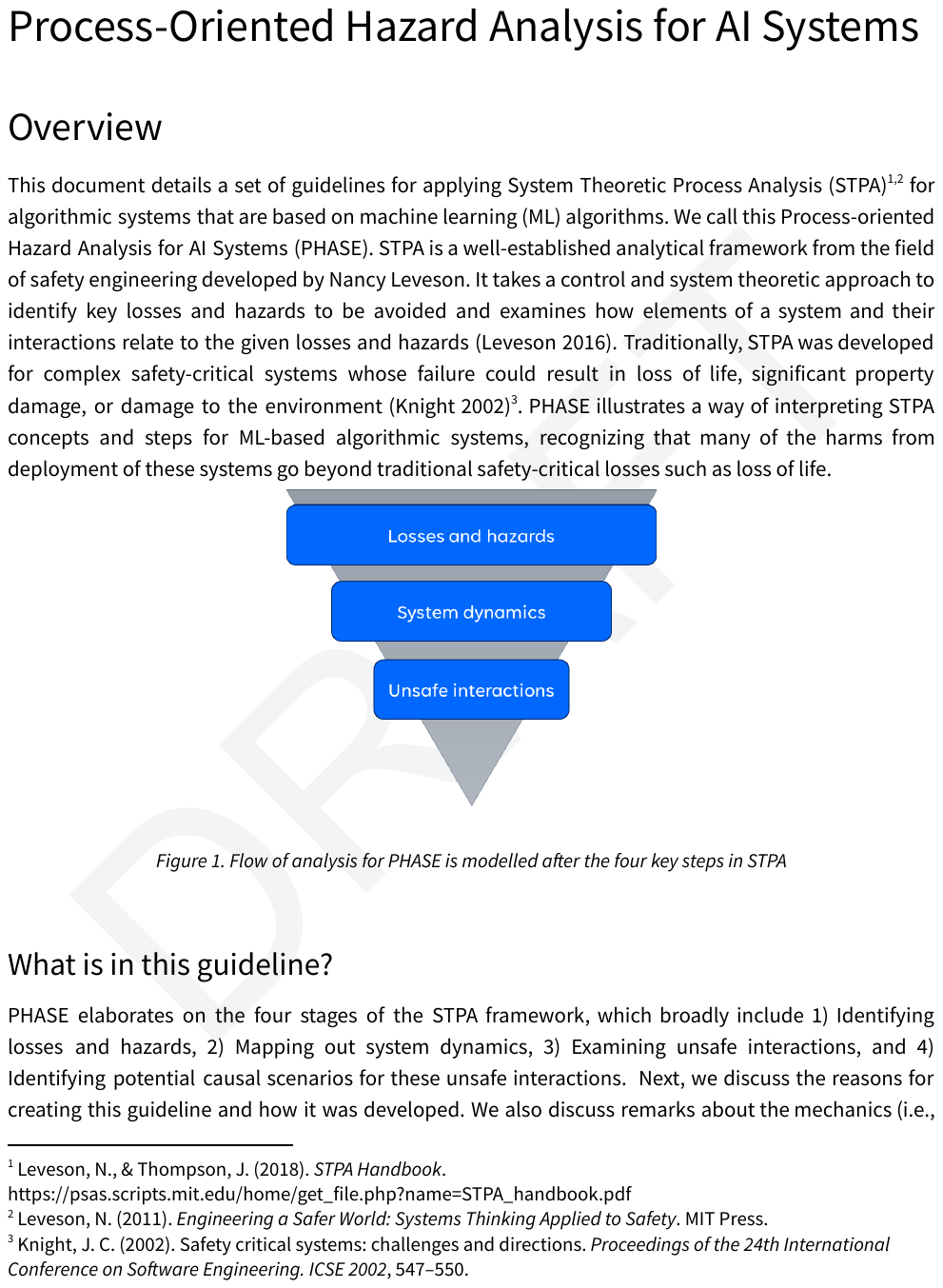}








\end{document}